\documentclass[11pt]{article}
\usepackage{CJKutf8}
\usepackage{booktabs}
\usepackage{stfloats}
\usepackage{booktabs} 

\usepackage[final]{acl}
\usepackage{multirow}
\usepackage{booktabs}
\usepackage{geometry}
\usepackage{colortbl}  
\usepackage{amssymb}
\usepackage{xcolor}
\definecolor{goodgreen}{HTML}{006400}

\usepackage{times}
\usepackage{latexsym}
\usepackage{enumitem}
\usepackage{amsmath}
\usepackage{booktabs}

\usepackage[T1]{fontenc}

\usepackage[utf8]{inputenc}

\usepackage{microtype}

\usepackage{inconsolata}

\usepackage{graphicx}

\title{Neuron-Guided Fine-Tuning: Unlocking Efficient Alignment Mechanisms for Large Language Models
}

\author{Zeyu Wu$^{1,}$\Thanks{Equal contribution.} ~~~ Junchao Wu$^{1,2,*}$ ~~~ Shudong Liu$^{1}$ ~~~ Runzhe Zhan$^{1}$ ~~~ Xin Chen$^{3}$ ~~~ Shu Yang$^{4}$ \\\bf Yichao Du$^{2,5}$ ~~~ \bf Longyue Wang$^{2}$ ~~~ \bf Weihua Luo$^{2}$ ~~~ \bf Jinsong Su$^{6}$ ~~~ \bf Derek F. Wong$^{1,}$\thanks{Corresponding author.} \\
$^{1}$NLP$^2$CT Lab, Faculty of Information Science and Computing, University of Macau\\
$^{2}$Alibaba Group ~~~ 
$^{3}$Nanjing University ~~~ 
$^{4}$KAUST ~~~ 
$^{5}$Wuhan University ~~~ 
$^{6}$Xiamen University\\
\texttt{nlp2ct.\{zeyu,junchao,shudong,runzhe\}@gmail.com, derekfw@um.edu.mo}\\
\texttt{x.chen@smail.nju.edu.cn, shu.yang@kaust.edu.sa, 
echaoustc@gmail.com}\\
\texttt{\{wanglongyue.wly,weihua.luowh\}@alibaba-inc.com, jssu@xmu.edu.cn}
}

\begin{document}
\maketitle
\begin{abstract}
Existing Supervised Fine-Tuning paradigms, particularly Full Parameter Fine-Tuning are often plagued by parameter redundancy, inconsistent data quality, and catastrophic forgetting, which current methods typically address in isolation and lack a unified optimization signal to bridge data selection, parameter updates, and knowledge preservation.
To address this, we propose \textbf{N}euron-\textbf{G}uided \textbf{F}ine-\textbf{T}uning \textbf{(\textsc{Ngft})}, a holistic framework that leverages neuron activation patterns as a universal proxy to unify the fine-tuning lifecycle. \textsc{Ngft} operates via three synergistic mechanisms: (1) Adaptive Task-Specific Neuron Selection, which identifies essential neurons in a single forward pass to concentrate updates and reduce redundancy; (2) Activation-Based Data Selection, which prioritizes information-dense samples that maximize contribution to key neurons; and (3) Neuron Activation Alignment, a novel loss function that anchors activations to pre-trained states, deepening representation learning and preserving general knowledge. 
Experimental results across three models across both domain-specific and general benchmarks demonstrate that \textsc{Ngft} significantly outperforms existing mainstream fine-tuning methods in both efficiency and performance, while effectively mitigating catastrophic forgetting.
Codes are publicly available at: \url{https://github.com/NLP2CT/NGFT}.
\end{abstract}

\section{Introduction}
Supervised Fine-Tuning (SFT) remains the primary approach to adapting Large Language Models (LLMs) from general pre-training to specific downstream tasks \cite{dong-etal-2024-abilities, hu-etal-2023-llm, harada-etal-2025-massive}. While Parameter-Efficient Fine-Tuning (PEFT) methods reduce memory overhead, Full Parameter Fine-Tuning (FPFT) retains its status as the golden standard for achieving strong performance \cite{lv-etal-2024-full, ding2022deltatuningcomprehensivestudy, NEURIPS2020_13b91943}. However, FPFT’s practical deployment is hindered by a fundamental bottleneck: the lack of a unified optimization signal that bridges data selection, parameter updates, and knowledge preservation across the fine-tuning lifecycle. This deficiency manifests in three intertwined challenges:
(1) Parameter Redundancy: LLM capabilities are encoded by sparse active neurons, yet FPFT forces updates on all neurons, causing computational resources waste and induces negative transfer \cite{xu2025let, leng-xiong-2025-towards, wang-etal-2024-expert,DBLP:journals/corr/abs-2503-23360}; 
(2) Inconsistent Data Quality: Instruction data often lack a universal utility selection criterion, which allows low-quality data to dilute the effective training signal and reduce data efficiency \cite{zhang2025survey, li-etal-2024-quantity};
(3) Catastrophic Forgetting: Domain-specific dense optimization overwrites pre-trained LLM general knowledge, reducing model versatility \cite{li-etal-2022-overcoming, huang-etal-2024-mitigating,DBLP:conf/aaai/WangWSLT20}.

Existing approaches typically address these issues in isolation, resulting in fragmented and heterogeneous strategies. Sparse fine-tuning \cite{xu2025let} targets parameter redundancy but ignores data quality, data selection \cite{wang-etal-2025-data-whisperer} improves data utility but decouples from parameter updates, and anti-forgetting techniques \cite{li-etal-2024-revisiting} are seldom integrated with the other two. As a result, fine-tuning still lacks a unified objective that jointly governs data selection, parameter optimization, and knowledge preservation.

To bridge this gap, we identify \textbf{Neuron Activation Patterns} as a universal proxy for fine-tuning. As a shared signal for data characteristics, parameter computation, and knowledge representation, they link model behavior to internal parameters and enable a unified framework for aforementioned three key dimensions in fine-tuning lifecycle.

Based on this, we propose Neuron-Guided Fine-Tuning (\textsc{Ngft}), a framework that utilizes neuron activation patterns as a universal proxy to construct a holistic full life-cycle optimization mechanism across three dimensions: (1) \textbf{Adaptive Task-Specific Neuron Selection}: We challenge the convention of fixed neuron selection ratios by addressing the heterogeneity of required knowledge across domains. We introduce a novel adaptive mechanism that identifies Task-Knowledge Neurons and dynamically calibrates optimal sparsity; (2) \textbf{Activation-Based Data Selection}: We pioneered a data selection criterion based on neuron activation intensity. By quantifying sample contributions to key neurons, we identify high-quality data located at the center of information density, improving the performance while reducing computational costs; (3) \textbf{Neuron Activation Alignment}: We propose a novel loss function for aligning neuron activation patterns. Introduced after cross-entropy loss stabilizes, it preserves general knowledge via pre-computed activation anchors while deepening task-specific representation learning, effectively mitigating catastrophic forgetting.

Experimental results on Llama-3.1 \cite{dubey2024llama}, Qwen2.5 \cite{qwen2.5}, and Mistral \cite{jiang2023mistral7b} across three domain-specific datasets (GSM8K \cite{cobbe2021training}, BioInstruct \cite{tran2024bioinstruct}, DialogSum \cite{chen2021dialogsum}), and three general benchmarks (MMLU \cite{hendrycks2020measuring}, BBH \cite{suzgun-etal-2023-challenging}, TyDiQA \cite{clark2020tydi}) demonstrate that \textsc{Ngft} consistently outperforms FPFT and mainstream fine-tuning frameworks. Averaged across all models, \textsc{Ngft} achieves a 4.66 point improvement in domain performance and mitigates forgetting-induced performance degradation by 9.89 point compared to FPFT. 
Further analysis shows that each component of \textsc{Ngft} remains effective when used independently, establishing strong baselines for different stages of the fine-tuning lifecycle.

\section{Related Work}

\subsection{Data Selection}
Data selection is pivotal in the instruction tuning of LLMs, aiming to extract high-quality subsets that enhance instruction-following capabilities while minimizing redundancy~\cite{zhang2025survey}. Existing mainstream selection paradigms primarily include: (1) Powerful LLM Scoring: leveraging powerful closed-source models (e.g., GPT-4) as external scorers to evaluate data quality and diversity \cite{li-etal-2024-quantity, li-etal-2024-one, chen2024alpagasus}; (2) Indicator-based Method: relying on heuristic metrics (e.g., syntactic complexity, perplexity) to measure the data diversity and importance~\cite{cao2023instruction, marion2023less, zheng2023coveragecentric}. (3) Gradient-based Approaches: quantifying data value by monitoring the gradient trajectories of the targeted model on specific data \cite{xia2024less, wang-etal-2025-data-whisperer}. However, these methods face inherent limitations: (1) the unreliability of indicator-based methods in high-dimensional spaces, and (2) the high cost of accessing closed-source APIs or gradient computation. While recent efforts explored neuron-aware data selection \cite{chen2026neuronaware}, these methods remain decoupled from parameter-level optimization and knowledge preservation, leaving the full potential of neuron activation patterns as a unifying signal untapped.

\subsection{Neuron Attribution and Fine-Tuning}
Feed-Forward Networks (FFNs), accounting for approximately two-thirds of LLM parameters, have emerged as a focal point in interpretability research. Widely regarded as the locus of knowledge storage, they are essential for encoding facts and integrating representations \cite{geva-etal-2021-transformer,meng2022locating, kim-etal-2025-unveiling}. Numerous neuron attribution studies aim to identify critical neurons within FFNs to gain deeper insights into LLM internals. Research on task-knowledge neurons primarily involves two paradigms: (1) Activation-based Attribution: localizing skill or language-specific neurons by analyzing activation offsets under soft prompts or cross-lingual patterns \cite{wang-etal-2022-finding-skill, tang2024language, christ-etal-2025-math}. (2) Gradient-based Attribution: employing techniques like integrated gradients to compute neuron-level contribution scores, pinpointing specific knowledge neurons \cite{dai2022knowledge, song-etal-2024-large, zhang2024unveiling}. The proven impact of these neurons on model behavior has inspired neuron-level parameter-efficient fine-tuning. For instance, \cite{xu2025let} demonstrated that updating only sensitive neurons in machine translation outperforms FPFT. Similarly, \cite{leng-xiong-2025-towards} showed that fine-tuning task-relevant neurons selected via gradient attribution improves performance while mitigating catastrophic forgetting in multi-task settings. However, existing neuron-based fine-tuning methods often lack adaptive mechanisms to dynamically adjust the ratio of key neurons based on specific domain or knowledge, thereby limiting model flexibility.

\section{Method}

\begin{figure*}[htbp]
    \centering
    \includegraphics[width=0.93\textwidth]{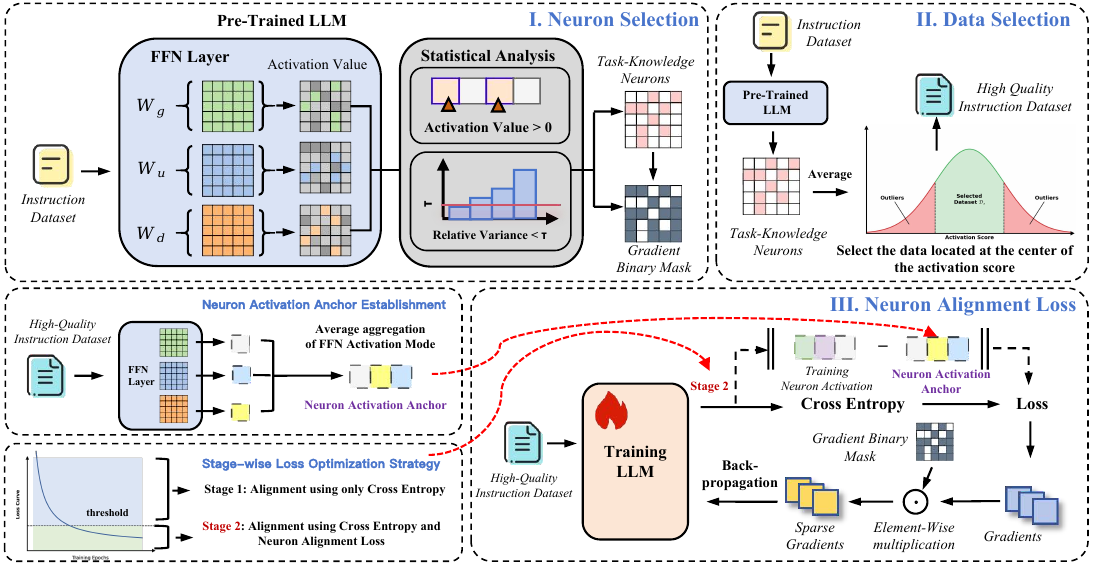}
    \caption{Overview of the proposed \textsc{Ngft} framework. It consists of three components: (I) Task-Knowledge Neuron Selection, which identifies sparse, stably activated neurons via coefficient of variation analysis; (II) Activation-Based Data Selection, which filters representative samples based on neuron activation intensity; and (III) Neuron-Level Alignment, which updates only critical neurons using gradient masking and incorporates a Neuron Alignment Loss to achieve deep alignment, thereby improving task performance while mitigating catastrophic forgetting.}
    \label{fig:NGFT_framework}
\end{figure*}

We propose an efficient neuron-guided fine-tuning framework named \textsc{Ngft}, illustrated in Figure~\ref{fig:NGFT_framework}, which consists of three components: (I) \textbf{Task-Knowledge Neuron Selection}: We adaptively localize a sparse set of neurons that are consistently active and stably activated across the entire training dataset. These neurons are regarded as the critical units with the most significant impact on model performance; (II) \textbf{Activation-Based Data Selection}: We select data samples whose activation patterns are more aligned with the distribution centroids of the task-knowledge neurons. These samples tend to best represent the core characteristics of the data; (III) \textbf{Neuron-Level
Alignment}: Once the cross-entropy (CE) loss reaches a specific convergence stage, we introduce a new objective that aligns the activation patterns of model responses with those induced by the ground truth. This guides the model to internalize deeper neuronal knowledge and achieve better representation alignment.

\subsection{Task-Knowledge Neuron Selection}

Mainstream LLMs (e.g., Llama 3, Qwen 2.5) predominantly adopt Transformer-based architectures that employ gated activation mechanisms within their FFNs. For the $l$-th layer, let the input hidden be denoted as $\tilde{\boldsymbol{h}}^i \in \mathbb{R}^d$. The computation of the FFN module output $\boldsymbol{h}^i$ is formalized as:
\begin{equation}
\boldsymbol{h}^i = \left( \sigma(\tilde{\boldsymbol{h}}^i W_{g}) \odot (\tilde{\boldsymbol{h}}^i W_{u}) \right) W_{d}
\end{equation}
where $W_g, W_u \in \mathbb{R}^{d \times 4d}$ and $W_d \in \mathbb{R}^{4d \times d}$ represent the Gate, Up, and Down projection matrices, respectively. $\sigma(\cdot)$ denotes the activation function, and $\odot$ represents element-wise multiplication. Structurally, we regard the column vectors of these projection matrices as neurons, and denote the activation of the $j$-th neuron in the $l$-th layer as $h_{l,j}$.

To identify Task-Knowledge Neurons, we define an instruction dataset $\mathcal{D} = \{x_1, x_2, \dots, x_m\}$, and a target LLM $\mathcal{M}$ with $L$ layers. We analyze the neuron activations of the final token of input $x_i$, which aggregates global semantic information \cite{zou2023representation, chen2025repreguard}. Based on this, we apply two constraints to identify Task-Knowledge Neurons: (1) Activation State, where a neuron is ``activated'' if its activation value for input $x_i$ is positive (i.e., $h_{l,j}^{(i)} > 0$), indicating an effective response to the input; (2) Distribution Stability, which filters out noisy neurons via the Coefficient of Variation. Neurons with low activation distribution Coefficient of Variation are stably activated and critical to model performance, consistent with \citet{wang-etal-2022-finding-skill, dai-etal-2022-knowledge} studies. Key neurons feature similarity and predictability. Specifically, for $h_{l,j}$, we calculate its mean activation value $\mu_{l,j}$ and standard deviation $\sigma_{l,j}$ across $\mathcal{D}$:

\begin{equation}
\mu_{l,j} = \frac{1}{m} \sum_{x_i \in \mathcal{D}} h_{l,j}^{(i)}
\end{equation}

\begin{equation}
\sigma_{l,j} = \sqrt{ \frac{1}{m - 1} \sum_{x_i \in \mathcal{D}} \left( h_{l,j}^{(i)} - \mu_{l,j} \right)^2 }
\end{equation}

We then obtain the Coefficient of Variation $R_{l,j} = \frac{\sigma_{l,j}}{\mu_{l,j} + \epsilon}$ for each neuron's distribution and identify a set of Task-Knowledge Neurons, denoted as $\mathcal{N}_k$. This subset comprises neurons $n_{l,j}$ that demonstrate high stability and satisfy the activation condition, determined by a threshold $\tau$:
\begin{equation}
\mathcal{N}_k = \{ n_{l,j} \mid R_{l,j} < \tau \text{ and } h_{l,j}^{(i)} > 0 \}.
\end{equation}

During fine-tuning, we apply gradient masking to the $\mathcal{N}_k$ to enable structured parameter updates while freezing non-essential neurons. Let ${\theta}$ represent the model parameters and $\mathcal{L}$ be the loss function. We construct a binary mask $\boldsymbol{M}$ in one-to-one corresponding to $\mathcal{N}_k$:
\begin{equation}
M_{l,j} = \begin{cases} 
1 & ,\ n_{l,j} \in \mathcal{N}_k \\ 
0 & ,\ \text{otherwise} 
\end{cases}
\end{equation}

In the backpropagation phase, constraints are imposed on the gradients via element-wise multiplication:
\begin{equation}
\nabla_{\boldsymbol{\theta}} \mathcal{L}^{\text{mask}} = \nabla_{\boldsymbol{\theta}} \mathcal{L} \odot \boldsymbol{M}
\end{equation}

This operation restricts updates to the parameters of key neurons, maintaining the pre-trained state of the remainder \cite{leng-xiong-2025-towards}. It effectively decouples precise fine-tuning from global parameter freezing, enabling the model to adapt to domain-specific data with significantly reduced computational costs.

\subsection{High-Quality Instruction Data Selection}

Upon identifying the Task-Knowledge Neurons $\mathcal{N}_k$, we evaluate each instruction instance by its activation profile on these neurons. To quantify this, we define the \textit{Task-Relevant Activation Score} for each instance $x_i$. Let $\mathcal{A}_{l,j}^{(i)}$ denote the activation value of neuron $n_{l,j} \in \mathcal{N}_k$ given input $x_i$. The score $S_i$ is:

\begin{equation}
S_i = \frac{1}{|\mathcal{N}_k|} \sum_{n_{l,j} \in \mathcal{N}_k} \mathcal{A}_{l,j}^{(i)}.
\end{equation}

However, selecting solely by magnitude can be misleading, as extreme activation spikes introduce skewness. Drawing on the information bottleneck framework and representation learning theory \cite{achille2018emergence, kawaguchi2023does}, instances proximal to the class centroid (mean value) are highly compressible yet lack representational diversity, whereas those distant from the centroid retain diversity but are disproportionately susceptible to corrupted samples. We thus propose a median-centered interval selection strategy that optimally resolves the information bottleneck trade-off
between two key criteria: Sufficiency (preserving task-relevant information) and Minimality (eliminating redundant noise). 

Specifically, let $r_i \in [r_1, \dots, r_m]$ denote the rank of instance $x_i$ sorted by $S_i$. We select a subset $\mathcal{D}_{s}$ based on a sampling ratio $\rho \in (0, 1]$:

\begin{equation}
\mathcal{D}_{s} = \left\{ x_i \in \mathcal{D} \mid \left| r_i - \frac{m}{2} \right| \le \frac{\rho \cdot m}{2} \right\}
\end{equation}

The continuous median neighborhood criterion naturally absorbs the core samples of all sub-clusters, thereby ensuring robustness to mild intra-class multimodal distribution while retaining the holistic distribution of the class \cite{xia2022moderate}.

\subsection{Neuron Alignment Loss Optimization}

While CE loss serves as the standard paradigm for SFT, it only optimizes the next-token probability distribution at the output layer. As gradient signals weaken during late convergence, models often hit performance plateaus. To address this, we propose a novel loss function based on neuron-level alignment. Focusing on internal model mechanisms, it guides deep knowledge learning and representation alignment in later optimization, easing convergence bottlenecks and enhancing model performance.

Formally, given a dataset $\mathcal{D}_s$ with $|\mathcal{D}_s| = n$, we sequentially feed samples into the target model $\mathcal{M}$ to capture domain-specific activation patterns. For each input $x_i$, we extract the activation vector $\omega_{pred}^{(i)}$ from the FFNs. This vector concatenates the neuron activations associated with the Gate, Up, and Down projection weights across layers. We then aggregate these vectors over $\mathcal{D}_s$ to construct the \textit{Neuron Activation Anchor}, denoted as $\mathbf{C}_{pred}$. This anchor characterizes the typical neuronal response pattern under the specific domain distribution:

\begin{equation}
    \mathbf{C}_{pred} = \frac{1}{n} \sum_{i=1}^{n} \omega_{pred}^{(i)}
\end{equation}

During the fine-tuning phase, let $\omega_{train}^{(i)}$ denote the real-time activation value of the FFNs given input $x_i$. By aligning the dynamic $\omega_{train}^{(i)}$ with the pre-computed activation anchor $\mathbf{C}_{pred}$, we explicitly guide the model to internalize domain-specific knowledge at a fine-grained neuron level. This alignment is achieved by minimizing the Mean Squared Error (MSE) between the current activation patterns and the anchor:
\begin{equation}
    \mathcal{L}_{\text{neu}} = \left\| \omega_{train}^{(i)} - \mathbf{C}_{pred} \right\|^2
\end{equation}

To balance training efficiency with model performance, we introduce an adaptive gating mechanism. Based on the hypothesis that incorporating Neuron Alignment Loss is most effective after the CE loss has stabilized, we utilize the CE loss $\mathcal{L}_{\text{ce}}$ as a proxy for training maturity. Specifically, the Neuron Alignment Loss $\mathcal{L}_{\text{neu}}$ is activated only when $\mathcal{L}_{\text{ce}}$ falls below a predefined threshold $\gamma$. The final objective combines both terms dynamically:
\begin{equation}
    \mathcal{L}_{\text{total}} = \mathcal{L}_{\text{ce}} + \mathbb{I}(\mathcal{L}_{\text{ce}} < \gamma) \cdot \mathcal{L}_{\text{neu}}
\end{equation}

This strategy ensures a coarse-to-fine optimization process, where the model first achieves precise semantic alignment before further exploring deep neural knowledge and representations.

\newcommand{\inc}[1]{\textcolor{green!70!black}{\scriptsize $\uparrow$#1\%}}
\newcommand{\dec}[1]{\textcolor{red}{\scriptsize $\downarrow$#1\%}}

\section{Experiments}
\subsection{Experimental Setup}
\paragraph{Instruction Tuning Datasets} Following \cite{wang-etal-2025-data-whisperer}, we evaluate model capabilities across distinct domains using three representative real-world instruction tuning datasets: GSM8K \cite{cobbe2021training} for mathematical reasoning, BioInstruct \cite{tran2024bioinstruct} for biomedical QA, and DialogSum \cite{chen2021dialogsum} for summarization. These datasets serve as benchmarks for both training and evaluation. We employ ROUGE-L to evaluate model performance on the BioInstruct and DialogSum datasets, while adopting Accuracy as the metric for the GSM8K. Detailed descriptions of the datasets can be found in Appendix \ref{sec:it_dataset_details}.

\paragraph{Datasets for Anti-Forgetting Testing} To assess the model's generalization ability and resistance to forgetting, we adopt MMLU \cite{hendrycks2020measuring}, BBH \cite{suzgun-etal-2023-challenging}, and TyDiQA \cite{clark2020tydi} as held-out benchmarks, aiming to verify whether the model retains its general-purpose knowledge question answering, reasoning, and multilingual capabilities after downstream task fine-tuning. For ease of subsequent analysis, we report the average score across these three benchmarks as the CF (Catastrophic Forgetting Evaluation) score, which quantifies the extent to which the model preserves its general capabilities throughout fine-tuning. Detailed descriptions of each dataset are provided in Appendix \ref{sec:additional_dataset_details}.

\paragraph{Models} To comprehensively evaluate the \textsc{Ngft} framework, we employ models from the Llama-3 \cite{dubey2024llama}, Qwen2.5 \cite{qwen2.5}, and Mistral \cite{jiang2023mistral7b} families. Our primary experiments utilize three representative models: Llama-3.1-8B-Instruct, Qwen2.5-7B-Instruct, and Mistral-7B-Instruct-v0.3. Furthermore, to verify the scalability and robustness, we also examine variants with diverse parameter sizes across these families (including 0.5B to 14B). Detailed results are provided in Appendix \ref{sec:model_details}.

\paragraph{Baselines}
To evaluate the effectiveness of our proposed framework, we perform comparative experiments from a modular perspective. In terms of data selection, we benchmark our method against random selection and established baselines such as Nuggets \cite{li-etal-2024-one}, CCS \cite{zheng2023coveragecentric}, LESS \cite{xia2024less}, and Data Whisperer \cite{wang-etal-2025-data-whisperer}. Regarding neuron selection, we compare our approach with random selection, MathNeuro \cite{christ-etal-2025-math}, NeFT \cite{xu2025let}, and NCFT \cite{leng-xiong-2025-towards}. For the loss function, we compare against the standard Cross-Entropy loss. Finally, at the framework level, we compare our method against FPFT (with random data selection) and a composite Strong Baseline (Data Whisperer + NCFT +  CE Loss). Detailed descriptions of these methods are provided in Appendix \ref{sec:baseline_details}.

\begin{table*}[!ht]
\centering
\renewcommand{\arraystretch}{0.88} 
\resizebox{0.95\textwidth}{!}{%
\begin{tabular}{llccccccccccccc}
\toprule
 & & & \multicolumn{4}{c}{\textbf{GSM8K}} & \multicolumn{4}{c}{\textbf{DialogSum}} & \multicolumn{4}{c}{\textbf{BioInstruct}} \\
\cmidrule(lr){4-7} \cmidrule(lr){8-11} \cmidrule(lr){12-15}
\textbf{Model} & \textbf{Method} & \textbf{Metric} & 1\% & 5\% & 10\% & Full & 1\% & 5\% & 10\% & Full & 1\% & 5\% & 10\% & Full \\
\midrule

\multirow{8}{*}{\rotatebox[origin=c]{90}{\text{Qwen2.5-7B-Instruct}}} 

 & \multirow{2}{*}{Zero-shot} 
   & \cellcolor{gray!10}Test & \multicolumn{4}{c}{\cellcolor{gray!10}42.91}& \multicolumn{4}{c}{\cellcolor{gray!10}23.27}& \multicolumn{4}{c}{\cellcolor{gray!10}30.68} \\
 & & CF   & \multicolumn{4}{c}{--}    & \multicolumn{4}{c}{68.62} & \multicolumn{4}{c}{--}    \\
 & \multirow{2}{*}{FPFT} 
   & \cellcolor{gray!10}Test & \cellcolor{gray!10}77.48& \cellcolor{gray!10}78.99& \cellcolor{gray!10}80.81& \cellcolor{gray!10}84.91& \cellcolor{gray!10}34.49& \cellcolor{gray!10}35.17& \cellcolor{gray!10}36.68& \cellcolor{gray!10}37.78& \cellcolor{gray!10}36.12& \cellcolor{gray!10}37.53& \cellcolor{gray!10}36.92& \cellcolor{gray!10}40.21 \\
 & & CF   & 57.24 & 56.25 & 56.31 & 58.09 & 65.96 & 64.33 & 64.33 & 57.88 & 62.18 & 65.85 & 62.30 & 58.10 \\

 & \multirow{2}{*}{Strong Baseline} 
   & \cellcolor{gray!10}Test & \cellcolor{gray!10}80.81& \cellcolor{gray!10}83.01& \cellcolor{gray!10}84.22& \cellcolor{gray!10}85.75& \cellcolor{gray!10}33.70& \cellcolor{gray!10}36.24& \cellcolor{gray!10}36.98& \cellcolor{gray!10}38.24& \cellcolor{gray!10}36.45& \cellcolor{gray!10}38.34& \cellcolor{gray!10}38.96& \cellcolor{gray!10}41.21 \\
 & & CF   & 65.46 & 65.35 & \underline{66.10} & \underline{67.20} & 68.81 & 67.81 & 67.95 & \underline{64.59} & 69.34 & 68.79 & 67.18 & 62.53 \\

 & \multirow{2}{*}{\textsc{Ngft}} 
   & \cellcolor{gray!10}Test & \cellcolor{gray!10}\textbf{82.18}& \cellcolor{gray!10}\textbf{84.61}& \cellcolor{gray!10}\textbf{86.13}& \cellcolor{gray!10}\textbf{87.64}& \cellcolor{gray!10}\textbf{34.83}& \cellcolor{gray!10}\textbf{37.15}& \cellcolor{gray!10}\textbf{38.16}& \cellcolor{gray!10}\textbf{39.41}& \cellcolor{gray!10}\textbf{38.79}& \cellcolor{gray!10}\textbf{39.60}& \cellcolor{gray!10}\textbf{40.97}& \cellcolor{gray!10}\textbf{42.27} \\
 & & CF  & \underline{65.75} & \underline{66.27} & 65.73 & 67.10 & \underline{69.99} & \underline{68.50} & \underline{69.34} & 64.21 & \underline{69.68} & \underline{70.31} & \underline{70.46} & \underline{65.34} \\

\midrule
\multirow{8}{*}{\rotatebox[origin=c]{90}{\text{Llama3.1-8B-Instruct}}} 
 & \multirow{2}{*}{Zero-shot} 
   & \cellcolor{gray!10}Test & \multicolumn{4}{c}{\cellcolor{gray!10}36.22}& \multicolumn{4}{c}{\cellcolor{gray!10}16.66}& \multicolumn{4}{c}{\cellcolor{gray!10}27.53} \\
 & & CF   & \multicolumn{4}{c}{--}    & \multicolumn{4}{c}{66.93} & \multicolumn{4}{c}{--}    \\
 & \multirow{2}{*}{FPFT} 
   & \cellcolor{gray!10}Test & \cellcolor{gray!10}62.77& \cellcolor{gray!10}67.37& \cellcolor{gray!10}69.67& \cellcolor{gray!10}77.71& \cellcolor{gray!10}32.77& \cellcolor{gray!10}34.77& \cellcolor{gray!10}35.64& \cellcolor{gray!10}38.10& \cellcolor{gray!10}30.70& \cellcolor{gray!10}33.79& \cellcolor{gray!10}34.82& \cellcolor{gray!10}40.57 \\
 & & CF   & 63.23 & 62.53 & 62.50 & 63.48 & 47.57 & 48.10 & 49.35 & 38.27 & 51.22 & 54.93 & 54.95 & 49.04  \\

 & \multirow{2}{*}{Strong Baseline} 
   & \cellcolor{gray!10}Test & \cellcolor{gray!10}61.11& \cellcolor{gray!10}73.10& \cellcolor{gray!10}75.51& \cellcolor{gray!10}77.64& \cellcolor{gray!10}34.84& \cellcolor{gray!10}\textbf{36.53}& \cellcolor{gray!10}37.27& \cellcolor{gray!10}38.58& \cellcolor{gray!10}34.34& \cellcolor{gray!10}35.83& \cellcolor{gray!10}36.63& \cellcolor{gray!10}40.84 \\
 & & CF   & \underline{68.15} & 67.80 & 68.56 & 68.25 & 56.21 & 58.22 & 58.47 & 48.30 & 62.16 & 61.85 & 60.03 & 55.05 \\

 & \multirow{2}{*}{\textsc{Ngft}} 
   & \cellcolor{gray!10}Test & \cellcolor{gray!10}\textbf{66.94}& \cellcolor{gray!10}\textbf{73.85}& \cellcolor{gray!10}\textbf{78.47}& \cellcolor{gray!10}\textbf{80.75}& \cellcolor{gray!10}\textbf{34.92}& \cellcolor{gray!10}36.33& \cellcolor{gray!10}\textbf{37.85}& \cellcolor{gray!10}\textbf{39.79}& \cellcolor{gray!10}\textbf{35.36}& \cellcolor{gray!10}\textbf{38.37}& \cellcolor{gray!10}\textbf{38.87}& \cellcolor{gray!10}\textbf{42.19} \\
 && CF  & 67.90 & \underline{68.61} & \underline{69.33} & \underline{68.86} & \underline{62.14} & \underline{60.62} & \underline{63.70} & \underline{54.43} & \underline{63.46} & \underline{63.98} & \underline{64.44} & \underline{56.80}  \\
 \midrule
 \multirow{8}{*}{\rotatebox[origin=c]{90}{\text{Mistral-7B-Instruct}}} 
 & \multirow{2}{*}{Zero-shot} 
   & \cellcolor{gray!10}Test & \multicolumn{4}{c}{\cellcolor{gray!10}25.24}& \multicolumn{4}{c}{\cellcolor{gray!10}20.46}& \multicolumn{4}{c}{\cellcolor{gray!10}21.40} \\
 & & CF   & \multicolumn{4}{c}{--}    & \multicolumn{4}{c}{60.56} & \multicolumn{4}{c}{--}    \\
 & \multirow{2}{*}{FPFT} 
   & \cellcolor{gray!10}Test & \cellcolor{gray!10}39.19& \cellcolor{gray!10}39.95& \cellcolor{gray!10}39.04& \cellcolor{gray!10}57.24& \cellcolor{gray!10}28.15& \cellcolor{gray!10}29.59& \cellcolor{gray!10}31.33& \cellcolor{gray!10}36.42& \cellcolor{gray!10}22.61& \cellcolor{gray!10}26.61& \cellcolor{gray!10}30.15& \cellcolor{gray!10}38.14 \\
 & & CF   & 51.26 & 50.70 & 49.88 & 47.20 & 26.20 & 18.12 & 15.54 & 14.02 & 25.18 & 28.75 & 26.33 & 12.80 \\

 & \multirow{2}{*}{Strong Baseline} 
   & \cellcolor{gray!10}Test & \cellcolor{gray!10}37.14& \cellcolor{gray!10}43.22& \cellcolor{gray!10}45.94& \cellcolor{gray!10}57.85& \cellcolor{gray!10}33.19& \cellcolor{gray!10}34.87& \cellcolor{gray!10}35.43& \cellcolor{gray!10}36.87& \cellcolor{gray!10}23.60& \cellcolor{gray!10}28.29& \cellcolor{gray!10}32.12& \cellcolor{gray!10}38.84 \\
 & & CF    & \underline{58.64} & 57.89 & 58.55 & 56.02 & 29.61 & 30.56 & 33.45 & 21.66 & 38.99 & 35.06 & 27.83 & 26.91 \\

 & \multirow{2}{*}{\textsc{Ngft}} 
   & \cellcolor{gray!10}Test & \cellcolor{gray!10}\textbf{42.46}& \cellcolor{gray!10}\textbf{47.61}& \cellcolor{gray!10}\textbf{52.01}& \cellcolor{gray!10}\textbf{62.55}& \cellcolor{gray!10}\textbf{33.96}& \cellcolor{gray!10}\textbf{34.95}& \cellcolor{gray!10}\textbf{36.51}& \cellcolor{gray!10}\textbf{38.06}& \cellcolor{gray!10}\textbf{28.50}& \cellcolor{gray!10}\textbf{32.42}& \cellcolor{gray!10}\textbf{33.99}& \cellcolor{gray!10}\textbf{39.60} \\
 & & CF   & 58.30 & \underline{58.29} & \underline{59.02} & \underline{58.07} & \underline{33.97} & \underline{32.17} & \underline{39.44} & \underline{27.71} & \underline{39.39} & \underline{47.08} & \underline{33.11} & \underline{30.77} \\
\bottomrule
\end{tabular}
}

\caption{Performance Comparison of \textsc{Ngft} with Zero-Shot, FPFT, and Strong Baseline (Data Whisperer + NCFT + CE Loss, represent the most SOTA baseline for data selection, neuron selection, and loss function optimization) when training with top 1\%, 5\%, 10\%, and full GSM8K, DialogSum, and BioInstruct data. The Test Metric measures in-domain performance, while the CF Metric assesses average performance on MMLU, BBH, and TyDiQA to reflect forgetting mitigation. \textbf{Bold} and \underline{underline} indicates best Test and CF performance.}
\label{tab:main_result}
\end{table*}

\subsection{Experimental Results}
\subsubsection{Main Results}

\paragraph{Superior Performance Across Datasets.} As shown in Table \ref{tab:main_result}, experimental results demonstrate that \textsc{Ngft} exhibits both robustness and efficiency across various domains and model architectures. In the full data setting, \textsc{Ngft} consistently outperforms FPFT on the Qwen2.5-7B-Instruct, Llama3.1-8B-Instruct, and Mistral-7B-Instruct-v0.3, achieving average gains of 2.14, 2.11, and 2.80 points across the three datasets, respectively. Furthermore, it surpasses Strong Baseline (Data Whisperer + NCFT + CE Loss) by average margins of 1.37, 1.89, and 2.21 points, respectively. Notably, \textsc{Ngft} excels in low-resource regimes. For examples, using Mistral-7B-Instruct-v0.3 on the GSM8K with 1\%, 5\%, and 10\% data rates, \textsc{Ngft} outperforms FPFT by 3.72, 7.66, and 12.97 points, respectively, while exceeding Strong Baseline by 5.32, 4.39, and 6.07 points. These results highlight its capability to extract high-quality features from sparse data. In addition, the cost analysis in Appendix \ref{sec:cost_time} demonstrates that \textsc{Ngft} can achieve performance comparable to FPFT with only 20.63\% of the computational cost. With an additional 11.3\% of the FPFT cost, it can consistently outperform FPFT by 1.5–3 points, validating the superior efficiency of the proposed framework.

\begin{table*}[!ht]
\centering
\renewcommand{\arraystretch}{0.88} 
\resizebox{\textwidth}{!}{%
\begin{tabular}{lcccccccccccc}
\toprule
 & \multicolumn{3}{c}{\textbf{GSM8K}} & \multicolumn{3}{c}{\textbf{DialogSum}} & \multicolumn{3}{c}{\textbf{BioInstruct}} \\
\cmidrule(lr){2-4} \cmidrule(lr){5-7} \cmidrule(lr){8-10}
\textbf{Method} & 1\% & 5\% & 10\% & 1\% & 5\% & 10\% & 1\% & 5\% & 10\% \\
\midrule
&\multicolumn{9}{c}{\textit{Qwen2.5-7B-Instruct}} \\
Random & 77.48 / 55.95 & 78.99 / 55.91 & 80.81 / 55.44  & 34.49 / 65.72 & 35.17 / 62.61 & 36.68 / 65.02 & 36.12 / 63.60 & 37.53 / 63.52 & 36.92 / 62.33  \\
CCS & 80.14 / 55.53 & 81.20 / \underline{56.31} & 82.11 / 56.89 & 34.44 / 66.21 & 36.00 / 63.28 & 36.74 / 64.63 & 37.23 / 64.02 & 37.99 / 62.87 & 38.51 / 61.68 \\
Nuggets& 78.47 / 55.44 & 80.51 / 55.34 & 82.87 / \underline{57.06} & 34.65 / 66.34 & 36.02 / 62.12 & 35.99 / 65.37 & 36.96 / 63.83 & 36.75 / 63.20 & 37.67 / 62.91\\
LESS & 80.14 / 55.89 & 81.05 / 56.11 & 82.03 / 56.27 & \textbf{34.96} / 65.98 & 35.78 / 63.99 & 36.62 / 65.12 & 36.88 / \underline{64.55} & 37.21 / 63.77 & 38.09 / 62.15 \\
Data Whisperer & 79.70 / 56.37  & 81.65 / 55.49 & 82.64 / 56.83 & 34.51 / 66.45 & \textbf{36.14} / 63.42 & 37.04 / 65.51 & 36.79 / 64.07 & 37.85 / 63.19 & 38.63 / 62.68 \\
$\textbf{\textsc{Ngft}}_{\text{DS}}$ & \textbf{80.41} / \underline{56.93} & \textbf{81.86} / 56.13 & \textbf{83.26} / 56.31 & 34.77 / \underline{67.21} & 36.05 / \underline{64.20} & \textbf{37.14} / \underline{66.28} & \textbf{38.35} / 63.39 & \textbf{38.60} / \underline{63.83} & \textbf{39.64} / \underline{63.32} \\
\midrule
&\multicolumn{9}{c}{\textit{Llama3.1-8B-Instruct}} \\
Random & 62.78 / 59.18 & 67.39 / 63.53 & 69.68 / 61.50 & 32.77 / \underline{49.15} & 34.71 / 45.29 & 35.64 / 49.15 & 30.70 / 47.83 & 33.79 / 55.83 & 34.82 / 51.80 \\
CCS & 62.47 / 61.68 & 70.81 / 62.81 & 72.79 / 62.47 & 34.08 / 48.95 & 35.19 / 45.96 & 36.35 / \underline{51.83} & 32.20 / 47.46 & 35.77 / \underline{56.40} & 36.59 / 52.82 \\
Nuggets& 63.00 / 60.26  & 71.65 / 62.95 & 72.33 / 62.41 & 33.00 / 47.93 & 35.46 / 46.09 & 35.55 / 50.17 & 31.41 / 48.14 & 34.12 / 54.25 & 35.60 / 53.27 \\
LESS & 62.85 / 60.91 & 71.57 / \underline{63.87} & \textbf{74.15} / 61.85 & 32.93 / 48.66 & \textbf{36.21} / 47.25 & 36.38 / 50.12 & 31.75 / 49.98 & 34.99 / 55.47 & 35.89 / \underline{54.36} \\
Data Whisperer & 63.14 / 61.46 & 71.27 / 63.19 & 73.54 / 62.43 & 33.48 / 48.09 & 35.53 / 46.71 & \textbf{36.73} / 50.88 & 32.27 / 49.51 & 34.66 / 56.13 & 36.47 / 53.84 \\
$\textbf{\textsc{Ngft}}_{\text{DS}}$ & \textbf{63.50} / \underline{62.60} & \textbf{72.32} / 63.60 & 73.26 / \underline{62.81} & \textbf{34.74} / 47.65 & 35.62 / \underline{47.57} & 36.64 / 50.03 & \textbf{34.02} / \underline{51.39} & \textbf{36.96} / 55.26 & \textbf{37.76} / 54.33 \\
\midrule
&\multicolumn{9}{c}{\textit{Mistral-7B-Instruct-v0.3}} \\
Random & 39.19 / 50.35 & 39.95 / 50.88 & 39.04 / 48.30 & 28.15 / 21.33 & 29.59 / 16.65 & 31.33 / 13.34 & 21.61 / 23.46 & 26.61 / 27.38 & 30.15 / 25.92 \\
CCS & \textbf{41.02} / 48.17 & 43.67 / 50.67 & 45.20 / 46.45 & 30.28 / 24.82 & 32.63 / 16.08 & 33.14 / 15.07 & 23.03 / 24.63 & 28.93 / 28.48 & 31.01 / 24.46 \\
Nuggets& 40.19 / 49.07 & 43.67 / 52.57 & 44.28 / 47.05 & 28.95 / 23.42 & 33.34 / 17.28 & 33.96 / 13.27 & 22.71 / \underline{25.33} & 28.37 / 27.98 & 29.91 / 25.46 \\
LESS & 40.48 / 49.31 & 43.29 / \underline{52.63} & 45.34 / 45.95 & 30.15 / 24.85 & 33.31 / 14.84 & 33.73 / 14.75 & 25.12 / 23.12 & 28.72 / \underline{29.48} & 31.69 / 25.68  \\
Data Whisperer& 40.80 / 49.82 & 43.89 / 52.18 & 45.79 / 46.68 & 30.19 / 22.24 & \textbf{33.52} / 15.22 & \textbf{34.28} / 14.46 & 24.85 / 23.61 & 29.53 / 28.95 & 31.27 / 26.34 \\
$\textbf{\textsc{Ngft}}_{\text{DS}}$ & 40.96 / \underline{51.00} & \textbf{43.98} / 51.03 & \textbf{45.85} / \underline{49.10} & \textbf{30.45} / \underline{25.18} & 33.38 / \underline{17.33} & 34.09 / \underline{15.14} & \textbf{25.42} / 24.96 & \textbf{30.32} / 28.55 & \textbf{32.62} / \underline{26.50} \\
\bottomrule
\end{tabular}
}
\caption{Performance (Test / CF) Comparison of \textsc{Ngft} with Data Selection Baselines when training with top 1\%, 5\%, and 10\% of the GSM8K, DialogSum, and BioInstruct data. The Test Metric measures in-domain performance, while the CF Metric assesses average performance on MMLU, BBH, and TyDiQA to reflect the model’s ability to mitigate forgetting. \textbf{Bold} indicates the best Test performance, and \underline{underline} indicates the best CF performance.}
\label{tab:table_2}
\end{table*}

\paragraph{Robust Resistance to Catastrophic Forgetting.} \textsc{Ngft} demonstrates exceptional efficacy in mitigating catastrophic forgetting and maintaining model generalization, even inducing significant positive transfer phenomena in certain settings. Specifically, experiments on Qwen2.5-7B-Instruct and Llama3.1-8B-Instruct show that, across varying proportions of the BioInstruct dataset, \textsc{Ngft} achieves average improvements of 6.84 points and 9.63 points compared to FPFT, and 1.98 points and 2.39 points over Strong Baseline. Notably, when fine-tuning Qwen2.5-7B-Instruct with only 5\% and 10\% of the BioInstruct data, the model's general capability scores (70.31 and 70.46) surpassed that of the Zero-shot (un-finetuned) baseline (68.62). This validates the superiority and practicality of the \textsc{Ngft} framework in preserving pre-trained knowledge while acquiring new domain expertise.

\subsubsection{Effectiveness of Adaptive Task-Knowledge Neuron Selection}
As showed in Table \ref{tab:table_3}, $\text{\textsc{Ngft}}_{\text{NS}}$ significantly surpasses baseline methods through its adaptive neuron selection. On Qwen2.5-7B-Instruct, it outperforms random selection by margins of 25.40, 8.89, and 9.84 points across three datasets. Moreover, compared to the SOTA method (NCFT), $\text{\textsc{Ngft}}_{\text{NS}}$ achieves consistent improvements in both task performance (up to 1.56 points) and knowledge retention (up to 3.86 points) across multiple model architectures. The key driver of this improvement is elucidated in Appendix \ref{sec:Adaptive Neuron Selection}. The adaptive activation variance threshold adopted by \textsc{Ngft} effectively resolves the inherent plasticity-stability dilemma in fixed-ratio methods, and the derived selection ratio generalizes well to other baselines. Moreover, as detailed in Appendix \ref{sec:neuron_selection_ratio}, ablation studies demonstrate that a Coefficient of Variation threshold within the range of 0.10 to 0.50 can consistently and stably improve performance. To ensure stability and experimental consistency, this work uniformly adopts the well-performing and stable value of 0.10 as the default Coefficient of Variation threshold across all experiments.

\begin{table}[!ht]
\centering
\renewcommand{\arraystretch}{0.88} 
\resizebox{0.4\textwidth}{!}{%
\begin{tabular}{lcccccccccccc}
\toprule
\textbf{Method} & \multicolumn{1}{c}{\textbf{GSM8K}} & \multicolumn{1}{c}{\textbf{DialogSum}} & \multicolumn{1}{c}{\textbf{BioInstruct}} \\
\midrule
&\multicolumn{3}{c}{\textit{Qwen2.5-7B-Instruct}} \\
Random & 62.17 / 63.31 & 29.87 / 63.08  & 32.12 / 62.90  \\
MathNeuro & 82.87 / 67.05 & 36.47 / 62.01 & 38.61 / 61.78 \\
NeFT & 83.57 / 65.09 & 37.93 / 61.84 & 40.47 / 62.52 \\
NCFT & 85.75 / 67.20 & 38.24 / 62.39 & 41.21 / 62.37  \\
$\textbf{\textit{\textsc{Ngft}}}_{\text{NS}}$ & \textbf{87.57} / \underline{67.28} & \textbf{38.76} / \underline{68.29} & \textbf{41.69} / \underline{66.68} \\
\midrule
&\multicolumn{3}{c}{\textit{Llama3.1-8B-Instruct}} \\
Random & 64.37 / 65.81 & 30.75 / 50.17 & 29.41 / 52.51 \\
MathNeuro & 75.97 / 66.70 & 37.48 / 49.34 & 36.61 / 52.27 \\
NeFT & 74.37 / 67.61 & 38.52 / 51.59 & 40.29 / 53.35 \\
NCFT & 77.64 / 66.25 & 38.58 / 48.30 & 40.84 / 55.05 \\
$\textbf{\textit{\textsc{Ngft}}}_{\text{NS}}$ & \textbf{79.91} / \underline{68.64} & \textbf{38.84} / \underline{55.98} & \textbf{41.46} / \underline{58.70} \\
\midrule
&\multicolumn{3}{c}{\textit{Mistral-7B-Instruct-v0.3}} \\
Random & 42.31 / 53.41 & 26.81 / 20.64 & 25.04 / 24.26 \\
MathNeuro & 53.30 / 53.26 & 35.69 / 26.85 & 34.23 / 25.51 \\
NeFT  & 56.71 / 52.92 & 36.41 / 25.62 & 37.58 / 23.17 \\
NCFT & 57.85 / 56.02 & 36.87 / 21.66 & 38.84 / 26.91 \\
$\textbf{\textit{\textsc{Ngft}}}_{\text{NS}}$ & \textbf{58.00} / \underline{56.02} & \textbf{37.30} / \underline{26.97} & \textbf{39.18} / \underline{29.50} \\
\bottomrule
\end{tabular}
}
\caption{Performance (Test / CF) Comparison of $\text{\textsc{Ngft}}_{\text{NS}}$ with Neuron Selection Baselines on full data.}
\label{tab:table_3}
\end{table}

\subsubsection{Effective Data Selection with Reduced Computational Costs}
As shown in Table \ref{tab:table_2}, the $\text{\textsc{Ngft}}_{\text{DS}}$ method exhibits superior performance. Compared to the SOTA baseline Data Whisperer, \textsc{Ngft} improves average performance on the BioInstruct dataset by 1.28, 1.78, and 0.90 points for Qwen2.5-7B-Instruct, Llama3.1-8B-Instruct, and Mistral-7B-Instruct-v0.3, respectively. Furthermore, models trained on subsets selected by \textsc{Ngft} consistently outperform other baselines in general capability metrics. While these gains are incremental, they serve as robust evidence of our method's capacity to mitigate catastrophic forgetting. Beyond performance, our approach offers a decisive advantage in computational efficiency. As detailed in Appendix \ref{sec:explore_data_selection_cost}, \textsc{Ngft} requires only a single forward pass with no auxiliary operations, achieving the lowest overhead among all compared methods. Notably, it delivers a 25.7$\times$ speedup over the Nuggets algorithm on the DialogSum dataset.
 
\subsubsection{Effectiveness and Intervention Timing of Neuron Alignment Loss}

\begin{figure}[htbp] 
    \centering
    \includegraphics[width=0.48\textwidth]{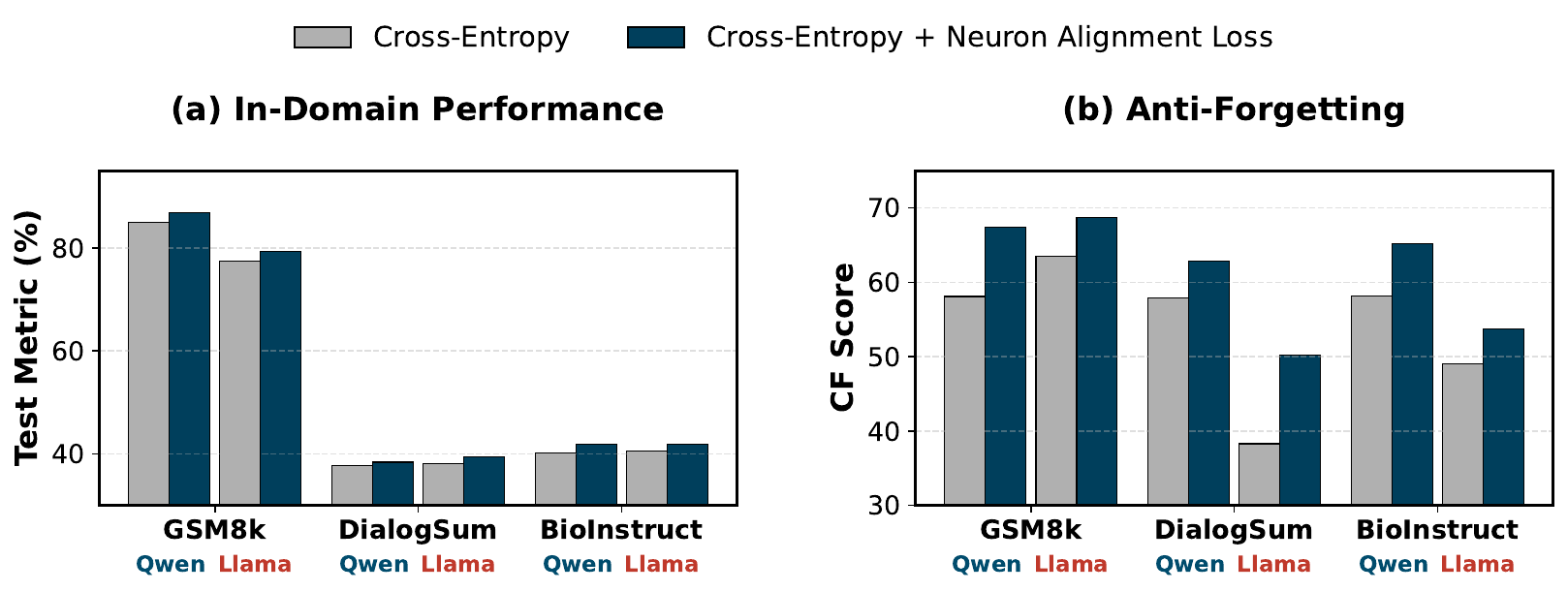}
    \caption{ Impact of Neuron Alignment Loss on model performance and ablation on intervention timing.}
    \label{fig:Impact of Neuron Alignment Loss}
\end{figure}

Figure \ref{fig:Impact of Neuron Alignment Loss} demonstrates the effectiveness of Neuron Alignment Loss On Qwen2.5-7B-Instruct and Llama3.1-8B-Instruct, in-domain performance improves by averages of 1.43 and 1.48 points, respectively, while general task performance increases by 7.11 and 7.29 points. These results indicate that Neuron Alignment Loss effectively captures neuron-level features to enhance domain capabilities and mitigate catastrophic forgetting during FPFT. Mechanistically, by anchoring initial activation patterns, Neuron Alignment Loss restricts the variation of non-critical neurons more strictly than CE loss. This effectively regularizes overfitting and achieves a superior trade-off between preserving prior knowledge and learning new tasks (see Appendix \ref{sec:explore_neuronloss} for detailed analysis). Furthermore, regarding the intervention timing, as shown in Appendix \ref{sec:Neuron Alignment Loss Intervention Timing}, introducing the neuron alignment loss when the CE loss converges to 0.03–0.10 can consistently improve the model performance. Based on ablation results, this work uniformly adopts the well-performing value of 0.04 as the intervention threshold across all experiments in this study.

\subsection{Analysis}
\begin{figure}[htbp]
    \centering
    \includegraphics[width=0.5\textwidth]{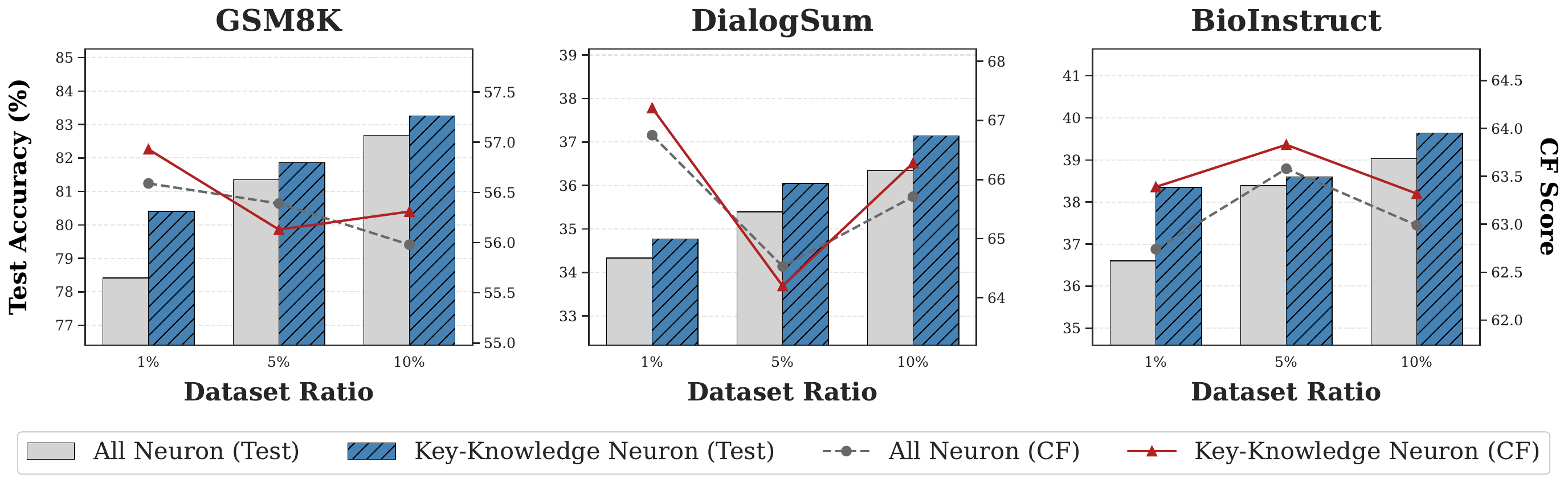}
    \caption{Impact of neuron analysis scope (all FFN neurons or Task-Knowledge Neurons) on data selection performance.}
    \label{fig:Impact of neuron analysis}
\end{figure}

\paragraph{Neuron Selection is Critical for Effective Data Selection.} 
We validate the necessity of data selection based on Task-Knowledge Neurons by comparing our strategy against selection based on the entire FFN. As shown in Figure \ref{fig:Impact of neuron analysis}, expanding the selection scope to the full FFN consistently degrades performance, resulting in drops of 1.03, 0.63, and 0.85 points on GSM8K, DialogSum, and BioInstruct, respectively. This decline is attributed to the noise introduced by the vast number of silent or randomly activated neurons within the full FFN. \textsc{Ngft} mitigates this by focusing exclusively on the Task-Knowledge Neuron subspace, effectively filtering noise and ensuring accurate data valuation.

\begin{figure}[htbp]
    \centering
    \includegraphics[width=0.5\textwidth]{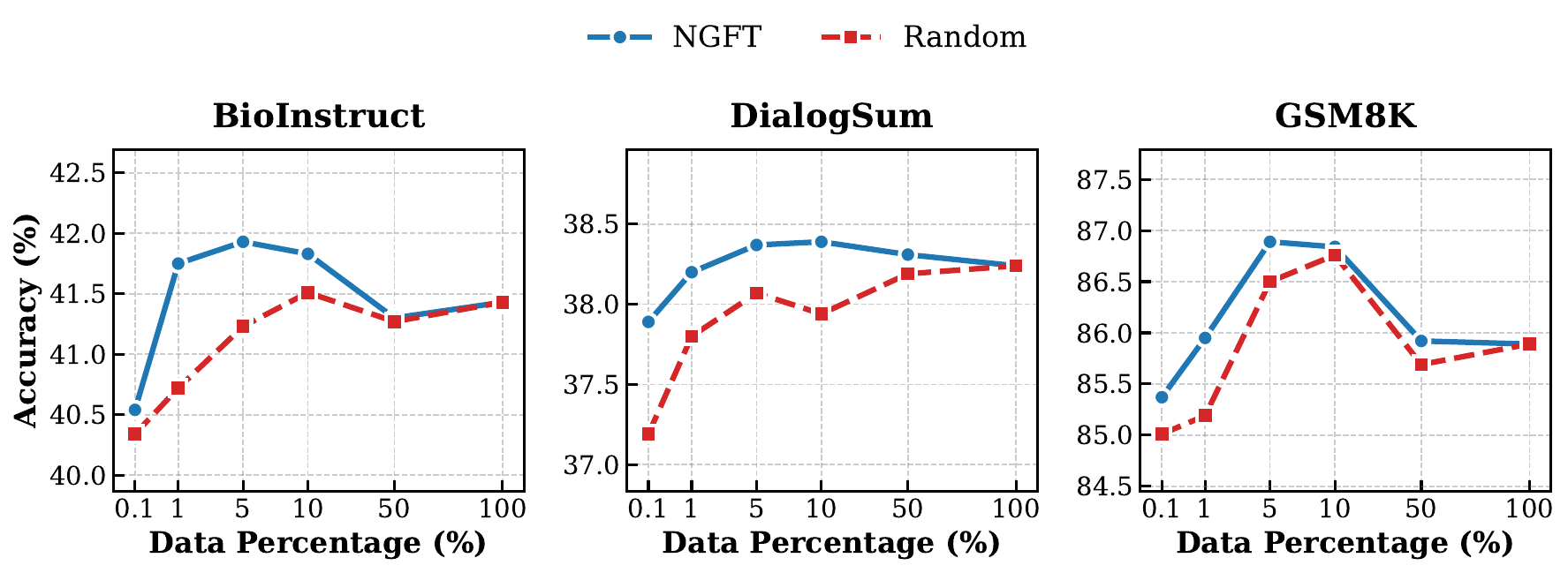}
    \caption{Comparison of anchor activation vector effectiveness for optimizing Neuron Alignment Loss across three data selection methods: random selection, \textsc{NGFT} selection, and the full dataset.}
    \label{fig:neuron_selection_and_loss}
\end{figure}

\paragraph{Data Selection Helps Optimize Neuron Alignment Loss.}  We evaluated three data selection strategies for extracting anchor activation vectors for neuron alignment loss: random sampling, \textsc{Ngft} framework, and full-data. As illustrated in Figure \ref{fig:neuron_selection_and_loss}, random sampling consistently underperforms compared to the full dataset. In contrast, \textsc{Ngft} demonstrates superior efficiency; utilizing only 5\% of the data selected by \textsc{Ngft} surpasses the full-data baseline. Interestingly, increasing this ratio to 50\% degrades performance, suggesting that an excessive inclusion of low-quality samples introduces noise and distorts the feature distribution. These results confirm that \textsc{Ngft} data selection successfully isolates high-value representative samples, ensuring superior neuron feature alignment with minimal data usage.

\paragraph{Scalability Across Different Model Sizes.} 

To assess the generalization and efficiency of \textsc{Ngft} on different scales, we extended our evaluation to other models in the Qwen2.5 series. As shown in Table \ref{tab:performance_compact_different_size_reformatted}, \textsc{Ngft} consistently achieves superior performance in all parameter sizes. For example, on Qwen2.5-14B-Instruct, it outperforms FPFT by average margins of 2.26 points in in-domain performance and 7.62 points in anti-forgetting capability across three datasets, and surpasses the strong baseline by 1.21 and 1.14 points, respectively. These results show that \textsc{Ngft} effectively adapts to diverse computational budgets and deployment scenarios, with exceptional scalability and robustness.

\section{Conclusion}
In this paper, we propose \textsc{Ngft}, a LLM fine-tuning framework guided by neuron activation patterns acting as a universal proxy. The framework integrates three modules: Task-Knowledge Neuron Selection, High-Quality Data Selection, and Neuron Alignment Loss Optimization, aiming to efficiently stimulate specific model capabilities for downstream tasks. Extensive experiments across three model families on both domain-specific and general benchmarks demonstrate that \textsc{Ngft} not only achieves superior in-domain performance compared to FPFT and strong baselines while effectively preventing catastrophic forgetting. The framework demonstrates excellent robustness and flexibility, offering a scalable solution for customizing LLMs in diverse scenarios.

\section*{Limitations}
While \textsc{Ngft} exhibits efficiency and robustness across models and benchmarks, our approach has inherent limitations. Grounded in neuron activation pattern analysis, \textsc{Ngft} requires access to model weights and internal states, making it applicable only to open-weights LLMs and inapplicable to proprietary black-box models such as GPT-5 with limited internal accessibility. Additionally, computational constraints precluded large-scale evaluations. With sufficient resources, future work will conduct larger-scale experiments to validate the generalization to more complex scenarios.

\section*{Acknowledgments}
This work was supported in part by the Science and Technology Development Fund of Macau SAR (Grant Nos. FDCT/0007/2024/AKP, EF2024-00185-FST), the UM and UMDF (Grant Nos. MYRG-GRG2024-00165-FST-UMDF, MYRG-GRG2025-00236-FST), the Tencent AI Lab Rhino-Bird Research Program (Grant No. EF2023-00151-FST), the Stanley Ho Medical Development Foundation (Grant No. SHMDF-AI/2026/001), and the National Natural Science Foundation of China (Grant No. 62266013).

\bibliography{custom}

\newpage

\appendix

\section{Dataset Details}
\subsection{Instruction Tuning Dataset Details}
\label{sec:it_dataset_details}

To comprehensively evaluate model performance, this experiment selects three instruction-tuning datasets with significant domain differences for training and testing. GSM8K \cite{cobbe2021training} serves as a high-quality benchmark for grade school math word problems, covering various topics ranging from basic arithmetic operations to fractions and percentages. The inclusion of this dataset aims to rigorously examine the mathematical logical reasoning capabilities of LLMs. BioInstruct \cite{tran2024bioinstruct} is an instruction-tuning dataset specifically customized for the biomedical domain. It contains 25,005 pairs of instructions and demonstration samples, encompassing a wide range of biomedical tasks. This dataset is utilized to verify the model's ability to process information and apply knowledge within specific medical fields. We split the dataset into training and test sets in a 9:1 ratio. DialogSum \cite{chen2021dialogsum} is a dialogue summarization dataset oriented toward real-life scenarios, involving diverse topics from daily life. This dataset is primarily used to test the model's comprehension of long texts as well as its capacity for extracting key information from dialogues and generating summaries. 

\subsection{Additional Dataset Details}
\label{sec:additional_dataset_details}

To validate the model's resistance to catastrophic forgetting and assess its generalization capabilities, we employ three additional datasets for evaluation. MMLU \cite{hendrycks2020measuring} serves as a comprehensive benchmark for evaluating multitask accuracy. It encompasses 57 diverse subjects, ranging from elementary to professional levels. Utilizing multiple-choice questions, MMLU is designed to rigorously assess the model's breadth of knowledge, cross-domain integration, and problem-solving abilities under both zero-shot and few-shot settings. BBH \cite{suzgun-etal-2023-challenging} is a curated subset of 23 challenging tasks selected from the BIG-Bench suite. It focuses on complex reasoning capabilities, such as symbolic manipulation, logical deduction, and commonsense reasoning.  TyDiQA \cite{clark2020tydi} is a multilingual question-answering benchmark covering 11 typologically diverse languages. This dataset evaluates the model's capacity for information retrieval and reading comprehension across diverse language families, making it critical for assessing multilingual generalization, particularly in non-English and low-resource contexts.

\subsection{Training Model Details}
\label{sec:model_details}
In this paper, we employed 8 distinct LLMs for training and evaluation. We designed three mainstream LLM frameworks covering a parameter range from 0.5B to 14B, specifically: Qwen2.5-0.5B-Instruct, Qwen2.5-3B-Instruct, Qwen2.5-7B-Instruct, Qwen2.5-14B-Instruct, Llama3.2-1B-Instruct, Llama3.2-3B-Instruct, Llama3.1-8B-Instruct, and Mistral-7B-Instruct-v0.3. 

All experiments were conducted using the LLAMA-Factory training framework on servers equipped with 8 NVIDIA A100 or H200 GPUs, leveraging DeepSpeed Zero-2.0 for training acceleration. To ensure fair comparison, we maintained consistent hyperparameter configurations across models for each task. Specifically, the learning rate was set to 2e-5 for all models on the DialogSum and BioInstruct datasets, while the learning rate was set to 1e-5 for the GSM8K dataset. Additionally, we utilized a batch size of 4 and a warmup ratio of 0.1 for all models. The training steps were uniformly set to 4 for all experiments. Detailed results for all models running within the \textsc{Ngft} framework can be found in Appendix \ref{sec:result_details}.

\subsection{Baseline Details}
\paragraph{Data Selection Baseline.} 
This paper selects four representative data selection methods as baselines for comparative study:

\begin{itemize}
    \item \textbf{Coverage-centric Coreset Selection} \cite{zheng2023coveragecentric}: The core principle of this method is to maximize coverage within the feature space. By analyzing the distribution of data features, it aims to identify the most representative subset of samples. The goal is to ensure this subset faithfully preserves the topological structure and diversity of the original dataset.

    \item \textbf{Nuggets} \cite{li-etal-2024-one}: This method leverages the few-shot learning capabilities of LLMs to explore the quality of instruction data. By constructing a multi-task predefined anchor set and calculating the impact of candidate samples on the perplexity of anchor set tasks, it generates a ``Golden Score'' to quantify sample quality.

    \item \textbf{LESS} \cite{xia2024less}: This is a data selection strategy based on gradient information. Utilizing low-rank gradient projection technology, this method calculates the influence of training data on specific target tasks, thereby precisely locating and selecting the high-quality data subset that contributes most significantly to model optimization.

    \item \textbf{Data Whisperer} \cite{wang-etal-2025-data-whisperer}: This method is based on the intrinsic connection between In-Context Learning and fine-tuning. By evaluating the contribution of samples when used as demonstrations and combining this with attention weight calibration, it achieves efficient data selection without the need for additional training.
\end{itemize}

\paragraph{Neuron Selection Baseline.} We conduct a comparative analysis using four representative neuron selection techniques as baselines.
\begin{itemize}
    \item \textbf{MathNeuro} \cite{christ-etal-2025-math}: This method employs the Wanda metric (defined as the product of the weight magnitude and the activation norm) to evaluate parameter importance across both mathematical and general domains. It identifies math-specific parameters by selecting the top-$K\%$ highest-scoring parameters on mathematical tasks and subsequently filtering out those that also rank within the top-$K\%$ for general tasks via a set difference operation.

    \item \textbf{NeFT} \cite{xu2025let}: This approach identifies key neurons by contrasting the pre-trained model with its fully fine-tuned counterpart. Specifically, it computes the cosine similarity between the weight vectors of corresponding neurons before and after fine-tuning. The top-$K\%$ neurons exhibiting the lowest similarity scores, which indicate the most significant directional shift in weights, are designated as task-sensitive neurons.

    \item \textbf{NCFT} \cite{leng-xiong-2025-towards}: This method adopts a gradient attribution technique to detect task-specific neurons. It approximates the impact of each neuron on the model's loss function using Taylor expansion to derive a relevance score. Based on these scores, the neurons are ranked, and the top-$k\%$ are selected as the key neurons.
\end{itemize}

\label{sec:baseline_details}

\section{Experiment Details}
\subsection{\textsc{Ngft} Adaptive Neuron Selection Precisely Enhances Model Performance.} 
\label{sec:Adaptive Neuron Selection}
In this section, we investigate the neuron retention ratios of the Qwen2.5-7B-Instruct, Llama3.1-8B-Instruct, and Mistral-7B-Instruct-v0.3 models under the \textsc{Ngft} framework, and we further explore the potential benefits of adaptive ratio strategies for other neuron selection methods. As shown in Table \ref{tab:model_performance_111111111111111}, we report the proportion of neurons selected by \textsc{Ngft}. Although the absolute number of activated Task-Knowledge neurons varies across datasets due to the heterogeneity of pre-training distributions, all three models exhibit a highly consistent trend: BioInstruct shows the lowest activation ratio, followed by DialogSum, with GSM8K demonstrating the highest. This phenomenon strongly validates the superior generalizability and robustness of the $\text{\textsc{Ngft}}_{\mathrm{NS}}$ method. Unlike fixed-ratio strategies, $\text{\textsc{Ngft}}_{\mathrm{NS}}$ dynamically adjusts the number of neurons based on the model's intrinsic assessment of task complexity and cognitive demands. This task-aware adaptive mechanism ensures an optimal trade-off between compression rate and performance across tasks of varying types and difficulties.

\begin{table}[htbp]
    \centering
    \resizebox{\columnwidth}{!}{
        \begin{tabular}{lccc}
            \toprule
            \textbf{Model}               & \textbf{BioInstruct} & \textbf{DialogSum} & \textbf{GSM8K} \\
            \midrule
            Qwen2.5-7B-Instruct         & 7.20\%               & 9.33\%             & 14.67\%        \\
            Llama3.1-8B-Instruct        & 5.28\%               & 8.23\%             & 12.18\%        \\
            Mistral-7B-Instruct-v0.3    & 1.19\%               & 14.42\%            & 17.03\%        \\
            \bottomrule
        \end{tabular}
    }
    \caption{Neuron selection ratios of $\textsc{Ngft}_{\mathrm{NS}}$ across different models and tasks.}
    \label{tab:model_performance_111111111111111}
\end{table}

In subsequent experiments, we incorporate the adaptive ratios identified by $\text{\textsc{Ngft}}_{\mathrm{NS}}$ into the strong baseline NCFT as prior knowledge. As shown in Table \ref{tab:model_performance_2222}, compared to conventional fixed ratios, these adaptive ratios effectively boosts the accuracy of NCFT across most scenarios, despite not outperforming the original $\text{\textsc{Ngft}}_{\mathrm{NS}}$. These results not only demonstrate the generalizability of the learned ratios but also further corroborate the superiority of \textsc{Ngft} in characterizing neuron importance distributions.

\begin{table}[htbp]
  \centering
  \resizebox{\columnwidth}{!}
  {\begin{tabular}{lccc}
    \toprule
    Model & \textbf{BioInstruct} & \textbf{DialogSum} & \textbf{GSM8K} \\
    \midrule
    Qwen2.5-7B-Instruct & 41.83 \small(\textcolor{goodgreen}{$\uparrow$0.61}) & 38.45 \small(\textcolor{goodgreen}{$\uparrow$0.21}) & 86.81 \small(\textcolor{goodgreen}{$\uparrow$1.05}) \\
    Llama3.1-8B-Instruct & 40.95 \small(\textcolor{goodgreen}{$\uparrow$0.11}) & 38.77 \small(\textcolor{goodgreen}{$\uparrow$0.19}) & 80.60 \small(\textcolor{goodgreen}{$\uparrow$2.96}) \\
    Mistral-7B-Instruct-v0.3 & 38.80 \small(\textcolor{red}{$\downarrow$0.04}) & 36.96 \small(\textcolor{goodgreen}{$\uparrow$0.09}) & 58.41 \small(\textcolor{goodgreen}{$\uparrow$0.56}) \\
    \bottomrule
  \end{tabular}
  }
  \caption{Experimental results of NCFT method on three models with neuron selection ratios from $\text{\textsc{Ngft}}_{\mathrm{NS}}$ using full dataset.}
  \label{tab:model_performance_2222}
\end{table}

\begin{table}[ht]
    \centering
    \small 
    \resizebox{0.43\textwidth}{!}{
        \begin{tabular}{lccc}
            \toprule
            \textbf{Method} & \textbf{GSM8K} & \textbf{DialogSum} & \textbf{BioInstruct} \\
            \midrule
            Nugget          & 1.73 & 2.83 & 2.35 \\
            CCS             & 0.93 & 1.16 & 1.08 \\
            LESS            & 1.54 & 2.07 & 2.21 \\
            Data Whisperer  & 0.41 & 0.64 & 0.81 \\
            \textsc{Ngft}   & 0.08 & 0.11 & 0.12 \\
            \midrule
            \rowcolor{blue!10}\textbf{Speedup}    & 21.6$\times$ & 25.72$\times$ &  19.58$\times$ \\
            \bottomrule
        \end{tabular}
    }
    \caption{STR metrics of different data selection methods on the Qwen2.5-7B-Instruct model. A lower STR indicates lower time complexity for the method.}
    \label{tab:time_performance}
\end{table}

\subsection{Comparison of Time Cost of Different Data Selection Methods.} 
\label{sec:explore_data_selection_cost}
Taking the Qwen2.5-7B-Instruct model as a case study, we further analyze the time overhead of different data selection methods. To quantitatively evaluate the efficiency of each method, we conducted data selection experiments using four A100 GPUs (simulating a resource-constrained setting) and employed the STR formula  \cite{wang-etal-2025-data-whisperer} for analysis. As shown in Table \ref{tab:time_performance}, our proposed method requires the least amount of time. Notably, on the DialogSum dataset, it achieves a 25.7$ \times$ speedup compared to the Nuggets, which incurs the highest computational cost. This is primarily because our method only requires a single forward pass for each data sample without the need for additional operations, thereby significantly reducing runtime.

\subsection{\textsc{Ngft} Outperforms Other PEFT Methods}
\label{sec:Comparison with Other PEFT Methods}
The motivation of this work is to pursue a more efficient FPFT paradigm. In the main experimental phase, we primarily conduct core performance comparisons against the advanced FPFT frameworks NCFT \cite{xu2025let} and NeFT \cite{leng-xiong-2025-towards}. To further comprehensively and systematically evaluate the overall performance and efficiency of \textsc{Ngft} framework, we additionally conduct extended comparisons with some Parameter-Efficient Fine-Tuning (PEFT) methods, specifically covering a suite of LoRA variants, including LoRA \cite{hu2022lora}, QLoRA \cite{dettmers2023qlora}, and DoRA \cite{liu2024dora}. All experiments are performed on the Qwen2.5-7B-Instruct model, and the full dataset is uniformly adopted for comparison, as these PEFT methods do not involve data selection. 

Detailed experimental results are presented in Table \ref{tab:comparison_peft}.
From the experimental results, we derive the following key findings:

(1) In terms of task performance, \textsc{Ngft} achieves state-of-the-art results across all benchmarks. On GSM8K, \textsc{Ngft} attains an accuracy of 87.64, representing an improvement of 1.74 points over the best-performing PEFT method DoRA, 4.24 points over LoRA, and 4.85 points over QLoRA. On DialogSum and BioInstruct, \textsc{Ngft} achieves ROUGE-L scores of 39.41 and 42.27, respectively, comprehensively surpassing all PEFT baselines. This demonstrates that the inherent constraints of low-rank fine-tuning restrict LoRA series methods from fully exploiting the model’s capabilities, whereas FPFT frameworks can unlock the full potential of the model within a larger optimization space.

(2) In terms of resistance to catastrophic forgetting, \textsc{Ngft} also exhibits highly competitive performance. FPFT performs drastically worse than all PEFT methods on the CF metric (only 58.09 on GSM8K), reflecting its inherent defect of severely impairing pre-trained knowledge in the absence of regularization constraints. By updating task-knowledge neurons and adopting the neuron alignment loss, \textsc{Ngft} effectively mitigates forgetting, achieving a CF score of 67.10 on GSM8K, which is marginally superior to the best PEFT method DoRA. It also exhibits the best catastrophic forgetting performance on DialogSum and BioInstruct.

(3) From the perspective of the comprehensive trade-off between performance and forgetting, the LoRA series delivers reasonable anti-forgetting performance owing to the structural prior of low-rank decomposition. However, restricting weight updates to a low-rank subspace inevitably limits the upper bound of task performance. Although DoRA partially alleviates this bottleneck via sparse selection and magnitude-direction decoupling, it remains constrained by the inherent limitations of parameter-efficient frameworks. In contrast, \textsc{Ngft} achieves state-of-the-art performance in both performance and anti-forgetting dimensions through neuron-level knowledge-aware gradient regulation, fully validating the necessity and superiority of the proposed method.

\begin{table}[htbp]
  \centering
  \small
  \resizebox{\linewidth}{!}{%
  \begin{tabular}{l cc cc cc}
    \toprule
    \multirow{2}{*}{\textbf{Method}} 
      & \multicolumn{2}{c}{\textbf{GSM8K}} 
      & \multicolumn{2}{c}{\textbf{DialogSum}} 
      & \multicolumn{2}{c}{\textbf{BioInstruct}} \\
    \cmidrule(lr){2-3} \cmidrule(lr){4-5} \cmidrule(lr){6-7}
      & Accuracy & CF & ROUGE-L & CF & ROUGE-L & CF \\
    \midrule
    Full Parameter & 84.91 & 58.09 & 37.78 & 57.88 & 40.21 & 58.10 \\
    LoRA           & 83.40 & 65.83 & 37.61 & 63.31 & 39.57 & 64.74 \\
    QLoRA          & 82.79 & 66.41 & 37.10 & 64.10 & 38.94 & 65.12 \\
    DoRA           & 85.90 & 66.47 & 38.25 & 63.73 & 41.15 & 65.01 \\
    $\text{\textsc{Ngft}}$ & \textbf{87.64} & \textbf{67.10} & \textbf{39.41} & \textbf{64.21} & \textbf{42.27} & \textbf{65.34} \\
    \bottomrule
  \end{tabular}
  }
  \caption{Performance comparison between \textsc{Ngft} and PEFT methods on Qwen2.5-7B-Instruct.}
  \label{tab:comparison_peft}
\end{table}

\begin{figure}[htbp]
    \centering
    \includegraphics[width=0.47\textwidth]{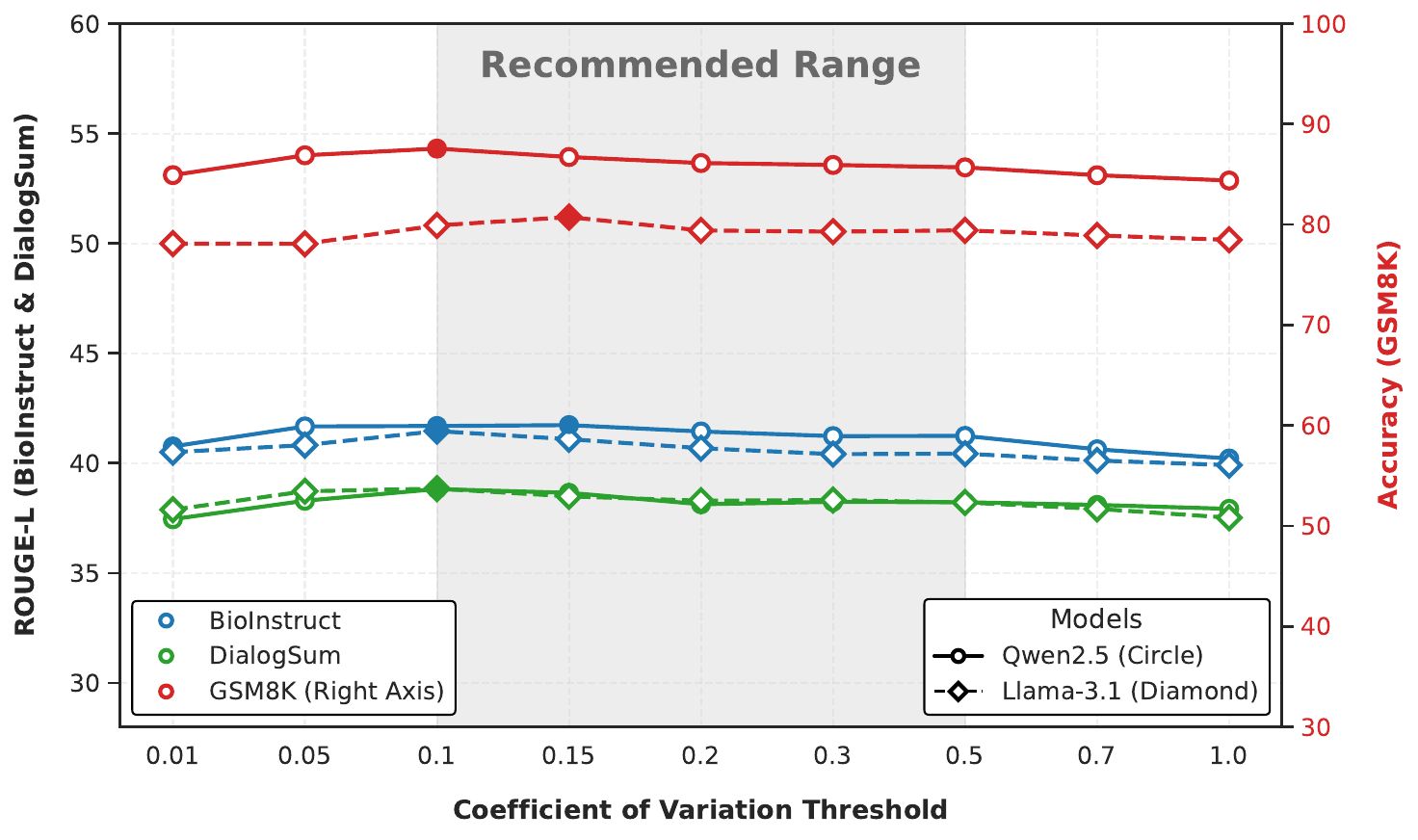}  
    \caption{Performance comparison of neurons selected by different Coefficient of Variation thresholds on Qwen2.5-7B-Instruct and Llama3.1-8B-Instruct.}
    \label{fig:neuron_ratio}
\end{figure}

\subsection{Performance Comparison of Different Coefficient of Variation Thresholds.}  
\label{sec:neuron_selection_ratio}
In \textsc{Ngft}, neuron selection is determined by a threshold derived from the coefficient of variation of activation distributions. To identify the optimal threshold, we perform ablation studies on Qwen2.5-7B-Instruct and Llama3.1-8B-Instruct across three diverse downstream tasks: GSM8K, DialogSum, and BioInstruct. Specifically, we evaluate coefficient of variation thresholds sampled from the set \{0.01, 0.05, 0.10, 0.15, 0.20, 0.30, 0.40, 0.50, 0.70, 1.00\}, with only the parameters of selected neurons updated during fine-tuning. 

Figure \ref{fig:neuron_ratio} illustrates a consistent trend across all experimental settings: as the coefficient of variation threshold increases, task performance first improves, plateaus, and ultimately degrades. This trajectory underscores an inherent trade-off between the granularity of neuron selection and model performance. Specifically, a conservative threshold (coefficient of variation < 0.10) yields an sparse number of selected neurons, which fails to adequately capture task-relevant features and thus constrains performance improvements. Conversely, an overly permissive threshold (coefficient of variation > 0.50) incorporates some task-irrelevant redundant neurons, thereby degrading model performance. 

Empirical results demonstrate that thresholds in the range of 0.10–0.50 consistently deliver stable performance gains, establishing this interval as a practical reference for real-world deployment. Optimal thresholds depend on model architecture and task characteristics: Llama3.1-8B-Instruct peaks at 0.15 (GSM8K) and 0.10 (DialogSum), while Qwen2.5-7B-Instruct performs best on GSM8K at 0.10. Thus, coefficient of variation thresholds are not universal, and task-specific tuning within 0.05–0.50 is recommended. Furthermore, from the perspective of mitigating catastrophic forgetting, a smaller coefficient of variation threshold correlates with fewer updated neurons, implying that smaller thresholds should be prioritized. Across all experimental configurations, a threshold of 0.10 consistently ranks among the top two performers, exhibiting strong robustness across both models and tasks. Consequently, to ensure reproducibility and experimental consistency, we adopt 0.10 as the recommended default coefficient of variation threshold for \textsc{Ngft} in this work.

\subsection{Comparison of Different Neuron Alignment Loss Intervention Timing.}
\label{sec:Neuron Alignment Loss Intervention Timing}
To investigate the impact of Neural Alignment Loss on model performance enhancement across diverse intervention timings, we conduct ablation studies using the Qwen2.5-7B-Instruct and Llama3.1-8B-Instruct models on three downstream tasks. Specifically, we evaluate the performance gains of the models by adjusting the intervention timing of Neural Alignment Loss at CE Loss values of \{0.02, 0.03, 0.04, 0.05, 0.06, 0.10, 0.20, 0.40\}.

\begin{figure}[htbp]
    \centering
    \includegraphics[width=0.47\textwidth]{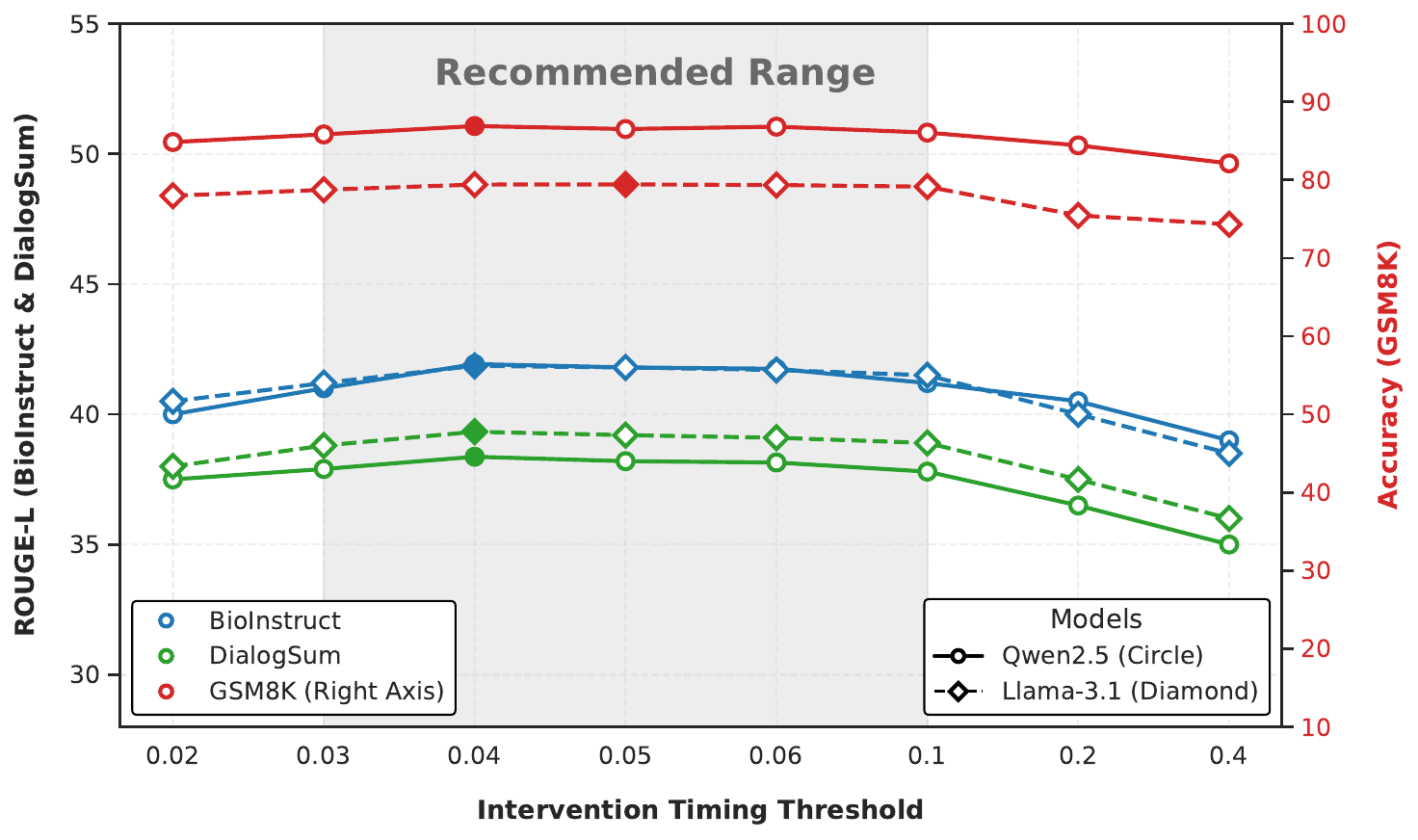}
    \caption{Performance comparison across different intervention timings of Neural Alignment Loss on Qwen2.5-7B-Instruct and Llama3.1-8B-Instruct.}
    \label{fig:ablation_threshold}
\end{figure}

Figure \ref{fig:ablation_threshold} reveals a consistent pattern across all settings: as the intervention timing of Neural Alignment Loss increases, task performance initially improves, then plateaus, and eventually degrades. This trend uncovers the distinct trade-offs of Neural Alignment Loss at different intervention points. Specifically, introducing Neural Alignment Loss too late (CE < 0.03) prevents the model from learning deep-level neuronal knowledge, thereby limiting the potential of performance improvement. Conversely, an overly early introduction (CE > 0.10) fails to enable effective alignment at the semantic level, which in turn weakens the model’s performance enhancement.

Empirical results demonstrate that the optimal threshold varies with model architectures and task characteristics. For instance, the optimal threshold for Qwen2.5-7B-Instruct on the GSM8K task is CE=0.03, while that for Llama3.1-8B-Instruct on the same task is CE=0.04. Furthermore, both models achieve sustained and stable performance improvements within the range of 0.03 to 0.10 with negligible performance gaps, making this interval a recommended reference for practical deployment. Across all experimental configurations, setting the threshold to CE=0.04 consistently yields superior model performance and robustness against catastrophic forgetting, demonstrating strong cross-model and cross-task generalization. Therefore, to ensure the reproducibility and consistency of experiments, this work uses CE=0.04 as the default intervention threshold for Neural Alignment Loss in this study.

\begin{table}[!ht]
\centering
\resizebox{0.5\textwidth}{!}{%
\begin{tabular}{lccc}
\toprule
\textbf{Section} & \multicolumn{1}{c}{\textbf{GSM8K}} & \multicolumn{1}{c}{\textbf{DialogSum}} & \multicolumn{1}{c}{\textbf{BioInstruct}} \\
\midrule
&\multicolumn{3}{c}{\textit{Only Cross-Entropy}} \\
Task-Knowledge Neurons & $2.98 \times 10^{-4}$ & $4.54 \times 10^{-4}$ & $6.81 \times 10^{-4}$ \\
Non-Task-Knowledge Neurons & $2.64 \times 10^{-4}$ & $4.22 \times 10^{-4}$ &  $6.67 \times 10^{-4}$\\
\midrule
&\multicolumn{3}{c}{\textit{Cross-Entropy + Neuron Alignment Loss}} \\
Task-Knowledge Neurons & $2.71 \times 10^{-4}$ & $4.26 \times 10^{-4}$ & $6.36 \times 10^{-4}$ \\
Non-Task-Knowledge Neurons & $1.89 \times 10^{-4}$ & $3.77 \times 10^{-4}$ & $4.48 \times 10^{-4}$ \\
\midrule
  $\Delta_\text{Task-Knowledge Neurons}$ & 9.06\% & 6.16\% & 6.60\% \\
 \rowcolor{blue!10} $\Delta_\text{Non-Task-Knowledge Neurons}$ & 28.40\% & 10.66\% & 32.83\% \\
\bottomrule
\end{tabular}
}
\caption{Comparison of changes in Task-Knowledge and Non-Task-Knowledge Neurons of Qwen2.5-7B-Instruct after fine-tuning with different loss functions.}
\label{tab:compare_neuron}
\end{table}

\subsection{Catastrophic Forgetting Induced by Using Neuron Alignment Loss} 
\label{sec:explore_neuronloss}
With Qwen2.5-7B-Instruct as our baseline model, we conducted a comparative analysis of only CE loss and CE loss augmented with Neuron Alignment Loss across three datasets, specifically focusing on their impact on internal neuronal representations during FPFT. As presented in Table \ref{tab:compare_neuron}, experimental results indicate that while the adjustment ratios for Task-Knowledge Neurons were comparable between the two methods, significant discrepancies emerged regarding Non-Task-Knowledge Neurons. Notably, on the BioInstruct dataset, this divergence reached 32.83\%. These findings suggest that Neuron Alignment Loss, by aligning with the activation patterns extracted from the initial model on the instruction dataset, effectively mitigates the drift of non-critical neurons during the later stages of fine-tuning, thereby equipping the model with robust resistance against catastrophic forgetting.

\begin{figure*}[htbp]
    \centering 
    \includegraphics[width=0.8\textwidth]{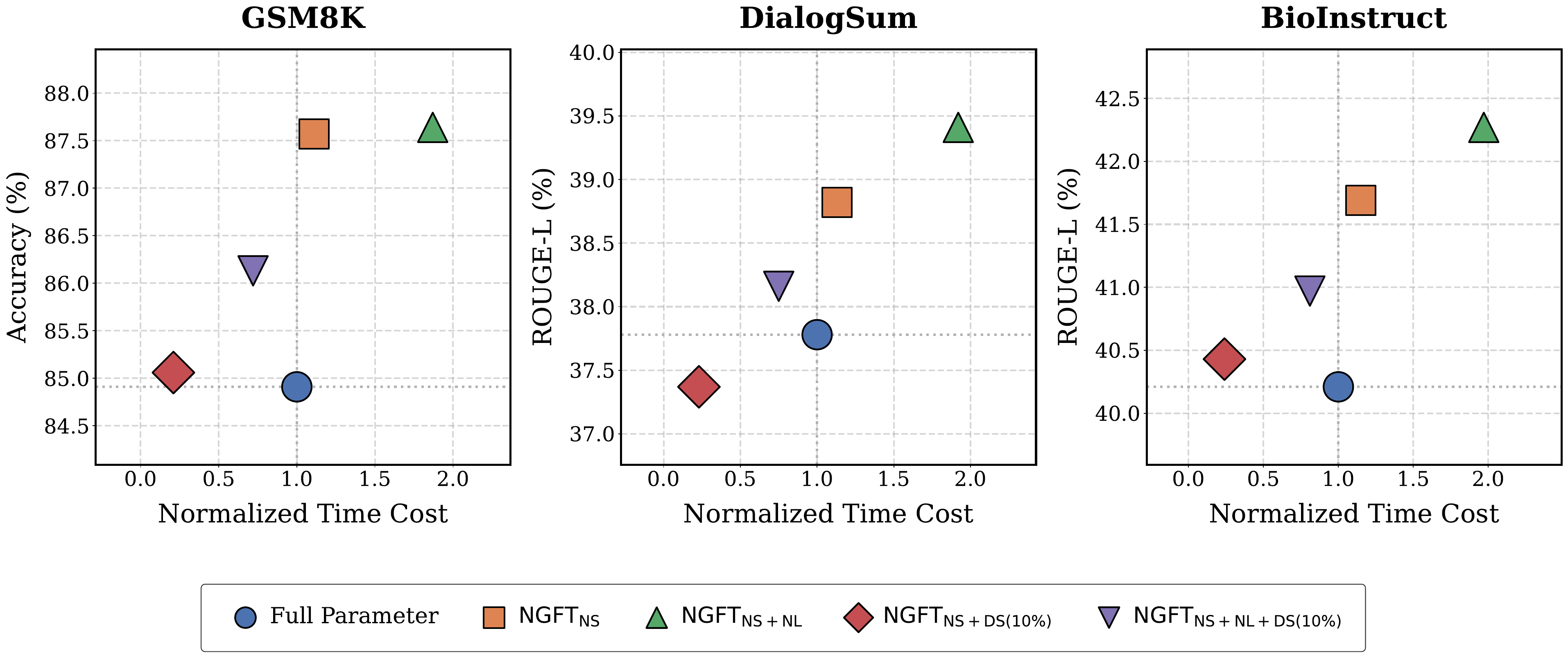}
    \caption{Analysis of In-Domain Performance and Time Overhead of the \textsc{Ngft} Framework on Qwen2.5-7B-Instruct Under Different Configurations. Specifically, $\text{\textsc{Ngft}}_{\text{NS}}$ fine-tunes only the Task-Knowledge Neurons; $\text{\textsc{Ngft}}_\text{NS+NL}$ incorporates a Neuron Alignment Loss on this basis; $\text{\textsc{Ngft}}_\text{NS+DS(10\%)}$ fine-tunes the aforementioned neurons using only 10\% of the selected data; while $\text{\textsc{Ngft}}_\text{NS+NL+DS(10\%)}$ combines both the 10\% data selection and the Neuron Alignment Loss.} 
    \label{fig:figure_6}
\end{figure*}

\subsection{Costs of Running of \textsc{Ngft}}
\label{sec:cost_time}
In this section, we analyze the time efficiency of \textsc{Ngft} using Qwen2.5-7B-Instruct as a representative case study. To account for variations in absolute training duration across different tasks, we introduce \textit{Normalized Time Cost} as a metric, defined as the ratio of a specific method's fine-tuning time to that of FPFT. This is calculated as:

\begin{equation}
    T_{\text{norm}} = \frac{T_{\text{method}}}{T_{\text{full}}}
\end{equation}

where $T_{\text{method}}$ denotes the training duration of the target method and $T_{\text{full}}$ denotes the duration of FPFT. 

As shown in Figure \ref{fig:figure_6} and Table \ref{tab:comparison_dif_comp_NGFT}, the experimental results demonstrate the dual advantages of \textsc{Ngft} under varying computational constraints:

First, in resource-constrained scenarios, \textsc{Ngft} exhibits superior computational efficiency. By synergizing Neuron Selection with Data Selection (utilizing only 10\% of the data), \textsc{Ngft} not only matches or even surpasses the performance of full-data FPFT but also reduces the average training time to merely 20.67\% of the latter, constituting a highly cost-effective fine-tuning baseline. Furthermore, while incorporating Neuron Alignment Loss introduces additional overhead due to auto-regressive generation, when combined with efficient data selection strategies, the total time cost remains controlled at 76.14\% of the full-data FPFT baseline, yet it consistently yields performance superior to that of FPFT. This indicates that high-quality data selection strategies effectively offset the temporal overhead incurred by the alignment loss, achieving an excellent balance between efficiency and performance.

Second, in performance-oriented scenarios, \textsc{Ngft} proves its capability to unleash the model's ultimate potential. Even with the full dataset, deploying the Neuron Selection module alone incurs a marginal average time overhead of 12.72\% for neuron localization, yet it consistently outperforms FPFT, demonstrating its robustness as an efficient fine-tuning solution. To further explore the performance upper bound, combining Neuron Alignment Loss increases the full-data training time to 92.31\% of the baseline; however, this configuration successfully achieves the maximum performance gain on in-domain tasks. This strategy of exchanging high computational investment for SOTA performance is well-suited for scenarios with ample computational budgets and strict requirements for model accuracy.

In summary, \textsc{Ngft} is not a rigid paradigm but a highly flexible, modular framework centered on neuron activation patterns. Each of its components (Neuron Selection, Data Selection, and Neuron Alignment Loss) has been independently verified to achieve advanced performance and features complete plug-and-play capability. This design empowers users to dynamically configure fine-tuning strategies according to specific computational resource budgets and performance metrics.

\begin{table}[htbp]
  \centering
  \small
  \resizebox{\linewidth}{!}{%
  \begin{tabular}{l c c c}
    \toprule
    \textbf{Method} & \textbf{GSM8K} & \textbf{DialogSum} & \textbf{BioInstruct} \\
    \midrule
    FPFT & 84.91 / 1.00 & 37.78 / 1.00 & 40.21 / 1.00 \\
    $\text{\textsc{Ngft}}_{\text{NS}}$ & 87.57 / 1.11 & 38.82 / 1.13 & 41.69 / 1.15 \\
    $\text{\textsc{Ngft}}_{\text{NS+NL}}$ & 87.64 / 1.87 & 39.41 / 1.92 & 42.27 / 1.97 \\
    $\text{NGFT}_{\text{NS+DS}(10\%)}$ & 85.06 / 0.21 & 37.37 / 0.20 & 40.43 / 0.21 \\
    $\text{\textsc{Ngft}}_{\text{NS+NL+DS}(10\%)}$ & 86.13 / 0.72 & 38.16 / 0.75 & 40.97 / 0.81 \\
    \bottomrule
  \end{tabular}
  }
\caption{Comparison of in-domain performance and time cost (Score / Normalized Time Cost) for Qwen2.5-7B-Instruct under Different \textsc{Ngft} Configurations.}
\label{tab:comparison_dif_comp_NGFT}
\end{table}

\subsection{In-Depth Analysis of Catastrophic Forgetting in \textsc{Ngft}}

This section details how FPFT induces performance degradation, specifically knowledge forgetting, style shifts, and position bias, and how the proposed \textsc{Ngft} framework resolves these issues.

(1) \textbf{Knowledge Forgetting and Negative Transfer in FPFT}:
Evaluation on mathematical tasks (using GSM8K) reveals that FPFT causes negative transfer in weakly correlated out-of-domain areas. While MMLU elementary mathematics and abstract algebra remain stable, college mathematics and statistics suffer average performance drops of 7.4, 8.6, and 9.3 points across the three evaluated LLMs. By forcing global adaptation to GSM8K's elementary reasoning style, FPFT overwrites the distinct activation patterns required for complex computational logic and data modeling in higher-level mathematics.

(2) \textbf{Mitigation and Positive Transfer via \textsc{Ngft}}:
\textsc{Ngft} effectively arrests this degradation through precise neuron regulation. Post-\textsc{Ngft}, the drop in college mathematics and statistics is constrained to a negligible 1.8 points. Furthermore, \textsc{Ngft} uniquely enables positive transfer by acquiring new skills while strictly preserving original knowledge structures: in elementary mathematics and abstract algebra, Qwen2.5-7B-Instruct (45.3 → 47.1) and Mistral-7B-v0.3 (34.7 → 35.9) demonstrate notable performance gains.

(3) \textbf{Disentangling Position Bias and Style Shifts}:
Beyond forgetting, FPFT triggers a severe problem-solving style shift. After FPFT on the biomedical BioInstruct dataset, models experience a catastrophic 50.7 point drop on MMLU college biology. Our analysis confirms this is driven by an extreme position bias, with the probability of choosing option "B" surging from 23.9\% to 71.5\%. This proves that unconstrained global updates can distort a model's foundational reasoning style~\cite{DBLP:conf/acl/ChenJZCYXYH26}. In contrast, \textsc{Ngft} actively resists this dataset-induced bias, limiting performance declines to within 2 points for two models (and a modest 5.3 points for Mistral-7B-Instruct), thereby maintaining the integrity of the model's problem-solving capabilities.

\section{Result Details}
\label{sec:result_details}
This section provides supplementary details on the experiments, specifically: (1) Table \ref{tab: detail_main} reports the detailed anti-forgetting performance of the models from Table 1 across three benchmarks: MMLU, BBH, and TyDiQA; (2) Table \ref{tab:performance_compact_different_size_reformatted} compares the performance of Qwen2.5 models of various sizes using \textsc{Ngft} against FPFT and strong baselines (Data Whisperer + NCFT + CE Loss); and (3) Tables \ref{tab: NGFT_0.5B} through \ref{tab:NGFT_8B_Llama} analyze the performance gains of eight models using individual and combined \textsc{Ngft} components, relative to zero-shot and full-data FPFT baselines.

\begin{table*}[htbp]
\centering
\renewcommand{\arraystretch}{0.84} 

\resizebox{\textwidth}{!}{%
\begin{tabular}{llccccccccccccc}
\toprule
 & & & \multicolumn{4}{c}{\textbf{GSM}} & \multicolumn{4}{c}{\textbf{DialogSum}} & \multicolumn{4}{c}{\textbf{BioInstruct}} \\
\cmidrule(lr){4-7} \cmidrule(lr){8-11} \cmidrule(lr){12-15}
\textbf{Model} & \textbf{Method} & \textbf{Metric} & 1\% & 5\% & 10\% & Full & 1\% & 5\% & 10\% & Full & 1\% & 5\% & 10\% & Full \\
\midrule

\multirow{16}{*}{\rotatebox[origin=c]{90}{\textbf{Qwen2.5-7B-Instruct}}} 

 & \multirow{4}{*}{Zero-shot} 
   & \cellcolor{gray!10}Test & \multicolumn{4}{c}{\cellcolor{gray!10}42.91}& \multicolumn{4}{c}{\cellcolor{gray!10}23.27}& \multicolumn{4}{c}{\cellcolor{gray!10}30.68} \\
 & & \small{MMLU}   & \multicolumn{4}{c}{--}    & \multicolumn{4}{c}{74.14} & \multicolumn{4}{c}{--}    \\
 & & \small{BBH}   & \multicolumn{4}{c}{--}    & \multicolumn{4}{c}{65.74} & \multicolumn{4}{c}{--}    \\
 & & \small{TyDiQA}   & \multicolumn{4}{c}{--}    & \multicolumn{4}{c}{65.99} & \multicolumn{4}{c}{--}    \\
 
 & \multirow{4}{*}{FPFT} 
   & \cellcolor{gray!10}Test & \cellcolor{gray!10}77.48& \cellcolor{gray!10}78.99& \cellcolor{gray!10}80.81& \cellcolor{gray!10}84.91& \cellcolor{gray!10}34.49& \cellcolor{gray!10}35.17& \cellcolor{gray!10}36.68& \cellcolor{gray!10}37.78& \cellcolor{gray!10}36.12& \cellcolor{gray!10}37.53& \cellcolor{gray!10}36.92& \cellcolor{gray!10}40.21 \\
 & & \small{MMLU}   & 65.31 & 63.31 & 62.81 & 66.19 & 70.81 & 70.38 & 70.15 & 65.16 & 67.41 & 67.58 & 66.28 & 62.74 \\
\bottomrule
\end{tabular}
}
\caption{Performance Comparison of \textsc{Ngft} with Zero-Shot, FPFT, and Strong Baselines (Data Whisperer + NCFT + CE Loss represent the SOTA baselines for data selection, neuron selection, and loss function optimization. ) when training with top 1\%, 5\%, 10\%, and full GSM8K, DialogSum, and BioInstruct data. }
\label{tab: detail_main}
\end{table*}

\begin{table*}[!ht]
\centering
\renewcommand{\arraystretch}{0.95} 
\resizebox{0.95\textwidth}{!}{%
\begin{tabular}{llccccccccccccc}
\toprule
 & & & \multicolumn{4}{c}{\textbf{GSM}} & \multicolumn{4}{c}{\textbf{DialogSum}} & \multicolumn{4}{c}{\textbf{BioInstruct}} \\
\cmidrule(lr){4-7} \cmidrule(lr){8-11} \cmidrule(lr){12-15}
\textbf{Model} & \textbf{Method} & \textbf{Metric} & 1\% & 5\% & 10\% & Full & 1\% & 5\% & 10\% & Full & 1\% & 5\% & 10\% & Full \\
\midrule

\multirow{8}{*}{\rotatebox[origin=c]{90}{\text{Qwen2.5-0.5B-Instruct}}} 

 & \multirow{2}{*}{Zero-shot} 
   & \cellcolor{gray!10}Test & \multicolumn{4}{c}{\cellcolor{gray!10}24.03}& \multicolumn{4}{c}{\cellcolor{gray!10}15.95}& \multicolumn{4}{c}{\cellcolor{gray!10}17.22} \\
 & & CF   & \multicolumn{4}{c}{--}    & \multicolumn{4}{c}{39.88} & \multicolumn{4}{c}{--}    \\

 & \multirow{2}{*}{FPFT} 
   & \cellcolor{gray!10}Test & \cellcolor{gray!10}31.18& \cellcolor{gray!10}32.67& \cellcolor{gray!10}35.92& \cellcolor{gray!10}39.30& \cellcolor{gray!10}28.95& \cellcolor{gray!10}30.72& \cellcolor{gray!10}31.58& \cellcolor{gray!10}34.88& \cellcolor{gray!10}28.39& \cellcolor{gray!10}29.68& \cellcolor{gray!10}31.71& \cellcolor{gray!10}36.40 \\
 & & CF   & 32.07 & 31.39 & 31.53 & 30.17 & 28.51 & 27.08 & 25.94 & 18.01 & 32.40 & 26.99 & 24.57 & 11.82 \\

 & \multirow{2}{*}{Strong Baseline} 
   & \cellcolor{gray!10}Test & \cellcolor{gray!10}40.96& \cellcolor{gray!10}42.79& \cellcolor{gray!10}44.55& \cellcolor{gray!10}47.56& \cellcolor{gray!10}32.46& \cellcolor{gray!10}33.70& \cellcolor{gray!10}34.00& \cellcolor{gray!10}36.40& \cellcolor{gray!10}32.24& \cellcolor{gray!10}34.37& \cellcolor{gray!10}34.78& \cellcolor{gray!10}39.10 \\
 & & CF   & 39.19 & 38.37 & 37.80 & 37.89 & 37.15 & 37.45 & 36.66 & 35.29 & 38.33 & 38.52 & 36.64 & 36.29 \\
\bottomrule
\end{tabular}
}
\caption{Performance (Test / CF) Comparison on Qwen2.5-0.5B-Instruct, Qwen2.5-3B-Instruct, and Qwen2.5-14B-Instruct. The Test Metric measures in-domain performance, while the CF Metric assesses average performance on MMLU, BBH, and TyDiQA to reflect the model’s ability to mitigate forgetting. \textbf{Bold} indicates the best Test performance, and \underline{underline} indicates the best CF performance.}
\label{tab:performance_compact_different_size_reformatted}
\end{table*}

\begin{table*}[htbp]
\centering
\resizebox{\textwidth}{!}{%
\begin{tabular}{llccccccccccccc}
\toprule
 & & & \multicolumn{4}{c}{\textbf{GSM}} & \multicolumn{4}{c}{\textbf{DialogSum}} & \multicolumn{4}{c}{\textbf{BioInstruct}} \\
\cmidrule(lr){4-7} \cmidrule(lr){8-11} \cmidrule(lr){12-15}
\textbf{Model} & \textbf{Method} & \textbf{Metric} & 1\% & 5\% & 10\% & Full & 1\% & 5\% & 10\% & Full & 1\% & 5\% & 10\% & Full \\
\midrule

\multirow{20}{*}{\rotatebox[origin=c]{90}{\textbf{Qwen2.5-0.5B-Instruct}}} 

 & \multirow{4}{*}{Zero-shot} 
   & \cellcolor{gray!10}Test & \multicolumn{4}{c}{\cellcolor{gray!10}24.03}& \multicolumn{4}{c}{\cellcolor{gray!10}15.95}& \multicolumn{4}{c}{\cellcolor{gray!10}17.22} \\
 & & \small{MMLU}   & \multicolumn{4}{c}{--}    & \multicolumn{4}{c}{46.87} & \multicolumn{4}{c}{--}    \\
 & & \small{BBH}   & \multicolumn{4}{c}{--}    & \multicolumn{4}{c}{32.31} & \multicolumn{4}{c}{--}    \\
 & & \small{TyDiQA}   & \multicolumn{4}{c}{--}    & \multicolumn{4}{c}{40.48} & \multicolumn{4}{c}{--}    \\

& \multirow{4}{4cm}{$\textsc{Ngft}_{\text{DS}} + \text{FPFT}$} 
   & \cellcolor{gray!10}Test & \cellcolor{gray!10}32.62& \cellcolor{gray!10}34.62& \cellcolor{gray!10}37.56& \cellcolor{gray!10}39.30& \cellcolor{gray!10}30.51& \cellcolor{gray!10}32.25& \cellcolor{gray!10}32.89& \cellcolor{gray!10}34.88& \cellcolor{gray!10}30.22& \cellcolor{gray!10}31.75& \cellcolor{gray!10}33.08& \cellcolor{gray!10}36.40 \\
 & & \small{MMLU}   & 40.83 & 40.12 & 40.21 & 40.22 & 40.41 & 38.83 & 36.84 & 33.02 & 38.98 & 37.42 & 35.49 & 12.87 \\
\bottomrule
\end{tabular}
}
\caption{Comparison of performance improvements on the Qwen2.5-0.5B-Instruct model using different \textsc{Ngft} components individually and in combination. }
\label{tab: NGFT_0.5B}
\end{table*}

\begin{table*}[htbp]
\centering
\label{tab:NGFT_3B_qwen}
\resizebox{\textwidth}{!}{%
\begin{tabular}{llccccccccccccc}
\toprule
 & & & \multicolumn{4}{c}{\textbf{GSM}} & \multicolumn{4}{c}{\textbf{DialogSum}} & \multicolumn{4}{c}{\textbf{BioInstruct}} \\
\cmidrule(lr){4-7} \cmidrule(lr){8-11} \cmidrule(lr){12-15}
\textbf{Model} & \textbf{Method} & \textbf{Metric} & 1\% & 5\% & 10\% & Full & 1\% & 5\% & 10\% & Full & 1\% & 5\% & 10\% & Full \\
\midrule
\multirow{20}{*}{\rotatebox[origin=c]{90}{\textbf{Qwen2.5-3B-Instruct}}} 
 & \multirow{4}{*}{Zero-shot} 
   & \cellcolor{gray!10}Test & \multicolumn{4}{c}{\cellcolor{gray!10}37.44}& \multicolumn{4}{c}{\cellcolor{gray!10}21.70}& \multicolumn{4}{c}{\cellcolor{gray!10}27.50} \\
  & & \small{MMLU}   & \multicolumn{4}{c}{--}    & \multicolumn{4}{c}{66.36} & \multicolumn{4}{c}{--}    \\
 & & \small{BBH}   & \multicolumn{4}{c}{--}    & \multicolumn{4}{c}{56.29} & \multicolumn{4}{c}{--}    \\
 & & \small{TyDiQA}   & \multicolumn{4}{c}{--}    & \multicolumn{4}{c}{55.44} & \multicolumn{4}{c}{--}    \\
 
 & \multirow{4}{4cm}{$\textsc{Ngft}_{\text{DS}} + \text{FPFT}$} 
   & \cellcolor{gray!10}Test & \cellcolor{gray!10}62.03& \cellcolor{gray!10}64.30& \cellcolor{gray!10}65.24& \cellcolor{gray!10}67.24& \cellcolor{gray!10}34.45& \cellcolor{gray!10}36.13& \cellcolor{gray!10}36.60& \cellcolor{gray!10}37.19& \cellcolor{gray!10}37.51& \cellcolor{gray!10}38.68& \cellcolor{gray!10}39.13& \cellcolor{gray!10}41.14 \\
 & & \small{MMLU}   & 61.35 & 61.99 & 61.81 & 58.06 & 61.71 & 61.61 & 61.13 & 58.33 & 61.32 & 59.90 & 60.36 & 57.43 \\
\bottomrule
\end{tabular}}
\caption{Comparison of performance improvements on the Qwen2.5-3B-Instruct model using different \textsc{Ngft} components individually and in combination.}
\end{table*}

\begin{table*}[htbp]
\centering
\label{tab:NGFT_7B_qwen}
\resizebox{\textwidth}{!}{%
\begin{tabular}{llccccccccccccc}
\toprule
 & & & \multicolumn{4}{c}{\textbf{GSM}} & \multicolumn{4}{c}{\textbf{DialogSum}} & \multicolumn{4}{c}{\textbf{BioInstruct}} \\
\cmidrule(lr){4-7} \cmidrule(lr){8-11} \cmidrule(lr){12-15}
\textbf{Model} & \textbf{Method} & \textbf{Metric} & 1\% & 5\% & 10\% & Full & 1\% & 5\% & 10\% & Full & 1\% & 5\% & 10\% & Full \\
\midrule
\multirow{20}{*}{\rotatebox[origin=c]{90}{\textbf{Qwen2.5-7B-Instruct}}} 
 & \multirow{4}{*}{Zero-shot} 
   & \cellcolor{gray!10}Test & \multicolumn{4}{c}{\cellcolor{gray!10}42.91}& \multicolumn{4}{c}{\cellcolor{gray!10}23.27}& \multicolumn{4}{c}{\cellcolor{gray!10}30.68} \\
  & & \small{MMLU}   & \multicolumn{4}{c}{--}    & \multicolumn{4}{c}{74.14} & \multicolumn{4}{c}{--}    \\
 & & \small{BBH}   & \multicolumn{4}{c}{--}    & \multicolumn{4}{c}{65.74} & \multicolumn{4}{c}{--}    \\
 & & \small{TyDiQA}   & \multicolumn{4}{c}{--}    & \multicolumn{4}{c}{65.99} & \multicolumn{4}{c}{--}    \\
 
 & \multirow{4}{4cm}{$\textsc{Ngft}_{\text{DS}} + \text{FPFT}$} 
   & \cellcolor{gray!10}Test & \cellcolor{gray!10}80.41& \cellcolor{gray!10}81.68& \cellcolor{gray!10}83.26& \cellcolor{gray!10}84.91& \cellcolor{gray!10}33.97& \cellcolor{gray!10}36.05& \cellcolor{gray!10}37.14& \cellcolor{gray!10}37.78& \cellcolor{gray!10}38.35& \cellcolor{gray!10}38.60& \cellcolor{gray!10}39.64& \cellcolor{gray!10}40.21 \\
 & & \small{MMLU}   & 66.04 & 63.02 & 63.23 & 66.19 & 71.05 & 70.61 & 70.33 & 65.16 & 68.52 & 68.24 & 67.08 & 62.74 \\
\bottomrule
\end{tabular}}
\caption{Comparison of performance improvements on the Qwen2.5-7B-Instruct model using different \textsc{Ngft} components individually and in combination.}
\end{table*}

\begin{table*}[htbp]
\centering
\label{tab:NGFT_14B_qwen}
\resizebox{\textwidth}{!}{%
\begin{tabular}{llccccccccccccc}
\toprule
 & & & \multicolumn{4}{c}{\textbf{GSM}} & \multicolumn{4}{c}{\textbf{DialogSum}} & \multicolumn{4}{c}{\textbf{BioInstruct}} \\
\cmidrule(lr){4-7} \cmidrule(lr){8-11} \cmidrule(lr){12-15}
\textbf{Model} & \textbf{Method} & \textbf{Metric} & 1\% & 5\% & 10\% & Full & 1\% & 5\% & 10\% & Full & 1\% & 5\% & 10\% & Full \\
\midrule
\multirow{20}{*}{\rotatebox[origin=c]{90}{\textbf{Qwen2.5-14B-Instruct}}} 
 & \multirow{4}{*}{Zero-shot} 
   & \cellcolor{gray!10}Test & \multicolumn{4}{c}{\cellcolor{gray!10}54.81}& \multicolumn{4}{c}{\cellcolor{gray!10}34.07}& \multicolumn{4}{c}{\cellcolor{gray!10}38.93} \\
  & & \small{MMLU}   & \multicolumn{4}{c}{--}    & \multicolumn{4}{c}{79.80} & \multicolumn{4}{c}{--}    \\
 & & \small{BBH}   & \multicolumn{4}{c}{--}    & \multicolumn{4}{c}{69.25} & \multicolumn{4}{c}{--}    \\
 & & \small{TyDiQA}   & \multicolumn{4}{c}{--}    & \multicolumn{4}{c}{67.48} & \multicolumn{4}{c}{--}    \\
 
 & \multirow{4}{4cm}{$\textsc{Ngft}_{\text{DS}} + \text{FPFT}$} 
   & \cellcolor{gray!10}Test & \cellcolor{gray!10}85.75& \cellcolor{gray!10}85.82& \cellcolor{gray!10}87.42& \cellcolor{gray!10}86.43& \cellcolor{gray!10}34.27& \cellcolor{gray!10}36.53& \cellcolor{gray!10}38.00& \cellcolor{gray!10}38.82& \cellcolor{gray!10}38.45& \cellcolor{gray!10}38.62& \cellcolor{gray!10}38.86& \cellcolor{gray!10}41.52 \\
 & & \small{MMLU}   & 72.10 & 69.05 & 69.30 & 72.25 & 79.90 & 79.45 & 79.10 & 73.81 & 77.11 & 76.47 & 74.88 & 69.12 \\
\bottomrule
\end{tabular}}
\caption{Comparison of performance improvements on the Qwen2.5-14B-Instruct model using different \textsc{Ngft} components individually and in combination.}
\end{table*}

\begin{table*}[htbp]
\centering
\label{tab:NGFT_7B_mistral}
\resizebox{\textwidth}{!}{%
\begin{tabular}{llccccccccccccc}
\toprule
 & & & \multicolumn{4}{c}{\textbf{GSM}} & \multicolumn{4}{c}{\textbf{DialogSum}} & \multicolumn{4}{c}{\textbf{BioInstruct}} \\
\cmidrule(lr){4-7} \cmidrule(lr){8-11} \cmidrule(lr){12-15}
\textbf{Model} & \textbf{Method} & \textbf{Metric} & 1\% & 5\% & 10\% & Full & 1\% & 5\% & 10\% & Full & 1\% & 5\% & 10\% & Full \\
\midrule
\multirow{20}{*}{\rotatebox[origin=c]{90}{\textbf{Mistral-7B-Instruct-v0.3}}} 
 & \multirow{4}{*}{Zero-shot} 
   & \cellcolor{gray!10}Test & \multicolumn{4}{c}{\cellcolor{gray!10}25.24}& \multicolumn{4}{c}{\cellcolor{gray!10}20.46}& \multicolumn{4}{c}{\cellcolor{gray!10}21.40} \\
  & & \small{MMLU}   & \multicolumn{4}{c}{--}    & \multicolumn{4}{c}{61.67} & \multicolumn{4}{c}{--}    \\
 & & \small{BBH}   & \multicolumn{4}{c}{--}    & \multicolumn{4}{c}{65.74} & \multicolumn{4}{c}{--}    \\
 & & \small{TyDiQA}   & \multicolumn{4}{c}{--}    & \multicolumn{4}{c}{54.29} & \multicolumn{4}{c}{--}    \\
 
 & \multirow{4}{4cm}{$\textsc{Ngft}_{\text{DS}} + \text{FPFT}$} 
   & \cellcolor{gray!10}Test & \cellcolor{gray!10}40.96& \cellcolor{gray!10}43.98& \cellcolor{gray!10}45.85& \cellcolor{gray!10}57.24& \cellcolor{gray!10}30.45& \cellcolor{gray!10}33.38& \cellcolor{gray!10}34.09& \cellcolor{gray!10}36.42& \cellcolor{gray!10}25.42& \cellcolor{gray!10}30.32& \cellcolor{gray!10}32.62& \cellcolor{gray!10}38.14 \\
 & & \small{MMLU}  & 57.09 & 56.24 & 54.47 & 52.98 & 35.82 & 26.45 & 26.32 & 25.03 & 35.84 & 27.56 & 29.51 & 25.08 \\
\bottomrule
\end{tabular}}
\caption{Comparison of performance improvements on the Mistral-7B-Instruct model using different \textsc{Ngft} components individually and in combination.}
\end{table*}

\begin{table*}[htbp]
\centering
\label{tab:NGFT_1B_llama}
\resizebox{\textwidth}{!}{%
\begin{tabular}{llccccccccccccc}
\toprule
 & & & \multicolumn{4}{c}{\textbf{GSM}} & \multicolumn{4}{c}{\textbf{DialogSum}} & \multicolumn{4}{c}{\textbf{BioInstruct}} \\
\cmidrule(lr){4-7} \cmidrule(lr){8-11} \cmidrule(lr){12-15}
\textbf{Model} & \textbf{Method} & \textbf{Metric} & 1\% & 5\% & 10\% & Full & 1\% & 5\% & 10\% & Full & 1\% & 5\% & 10\% & Full \\
\midrule
\multirow{20}{*}{\rotatebox[origin=c]{90}{\textbf{Llama3.2-1B-Instruct}}} 
 & \multirow{4}{*}{Zero-shot} 
   & \cellcolor{gray!10}Test & \multicolumn{4}{c}{\cellcolor{gray!10}12.51}& \multicolumn{4}{c}{\cellcolor{gray!10}14.02}& \multicolumn{4}{c}{\cellcolor{gray!10}22.59} \\
  & & \small{MMLU}   & \multicolumn{4}{c}{--}    & \multicolumn{4}{c}{46.00} & \multicolumn{4}{c}{--}    \\
 & & \small{BBH}   & \multicolumn{4}{c}{--}    & \multicolumn{4}{c}{38.42} & \multicolumn{4}{c}{--}    \\
 & & \small{TyDiQA}   & \multicolumn{4}{c}{--}    & \multicolumn{4}{c}{43.69} & \multicolumn{4}{c}{--}    \\
 
 & \multirow{4}{4cm}{$\textsc{Ngft}_{\text{DS}} + \text{FPFT}$} 
   & \cellcolor{gray!10}Test & \cellcolor{gray!10}32.15& \cellcolor{gray!10}33.20& \cellcolor{gray!10}34.41& \cellcolor{gray!10}38.66& \cellcolor{gray!10}31.70& \cellcolor{gray!10}33.32& \cellcolor{gray!10}34.52& \cellcolor{gray!10}36.92& \cellcolor{gray!10}31.96& \cellcolor{gray!10}34.63& \cellcolor{gray!10}35.52& \cellcolor{gray!10}38.50 \\
 & & \small{MMLU}   & 43.24 & 43.17 & 42.87 & 38.86 & 41.97 & 41.41 & 38.33 & 32.54 & 42.08 & 41.80 & 42.69 & 32.75 \\
\bottomrule
\end{tabular}}
\caption{Comparison of performance improvements on the Llama3.2-1B-Instruct model using different \textsc{Ngft} components individually and in combination. }
\end{table*}

\begin{table*}[htbp]
\centering
\label{tab:performance_optimized_filled}
\resizebox{\textwidth}{!}{%
\begin{tabular}{llccccccccccccc}
\toprule
 & & & \multicolumn{4}{c}{\textbf{GSM}} & \multicolumn{4}{c}{\textbf{DialogSum}} & \multicolumn{4}{c}{\textbf{BioInstruct}} \\
\cmidrule(lr){4-7} \cmidrule(lr){8-11} \cmidrule(lr){12-15}
\textbf{Model} & \textbf{Method} & \textbf{Metric} & 1\% & 5\% & 10\% & Full & 1\% & 5\% & 10\% & Full & 1\% & 5\% & 10\% & Full \\
\midrule
\multirow{20}{*}{\rotatebox[origin=c]{90}{\textbf{Llama3.2-3B-Instruct}}} 
 & \multirow{4}{*}{Zero-shot} 
   & \cellcolor{gray!10}Test & \multicolumn{4}{c}{\cellcolor{gray!10}29.45}& \multicolumn{4}{c}{\cellcolor{gray!10}15.12}& \multicolumn{4}{c}{\cellcolor{gray!10}26.88} \\
  & & \small{MMLU}   & \multicolumn{4}{c}{--}    & \multicolumn{4}{c}{60.45} & \multicolumn{4}{c}{--}    \\
 & & \small{BBH}   & \multicolumn{4}{c}{--}    & \multicolumn{4}{c}{58.61} & \multicolumn{4}{c}{--}    \\
 & & \small{TyDiQA}   & \multicolumn{4}{c}{--}    & \multicolumn{4}{c}{62.50} & \multicolumn{4}{c}{--}    \\
 
 & \multirow{4}{4cm}{$\textsc{Ngft}_{\text{DS}} + \text{FPFT}$} 
   & \cellcolor{gray!10}Test & \cellcolor{gray!10}62.25& \cellcolor{gray!10}67.40& \cellcolor{gray!10}68.46& \cellcolor{gray!10}70.97& \cellcolor{gray!10}33.31& \cellcolor{gray!10}36.13& \cellcolor{gray!10}37.09& \cellcolor{gray!10}38.21& \cellcolor{gray!10}35.47& \cellcolor{gray!10}37.56& \cellcolor{gray!10}38.53& \cellcolor{gray!10}41.49 \\
 & & \small{MMLU}   & 59.45 & 59.80 & 60.10 & 58.90 & 60.39 & 59.28 & 58.34 & 54.62 & 60.09 & 58.66 & 58.63 & 53.43 \\
\bottomrule
\end{tabular}}
\caption{Comparison of performance improvements on the Llama3.2-3B-Instruct model using different \textsc{Ngft} components individually and in combination.}
\end{table*}

\begin{table*}[htbp]
\centering
\resizebox{\textwidth}{!}{%
\begin{tabular}{llccccccccccccc}
\toprule
 & & & \multicolumn{4}{c}{\textbf{GSM}} & \multicolumn{4}{c}{\textbf{DialogSum}} & \multicolumn{4}{c}{\textbf{BioInstruct}} \\
\cmidrule(lr){4-7} \cmidrule(lr){8-11} \cmidrule(lr){12-15}
\textbf{Model} & \textbf{Method} & \textbf{Metric} & 1\% & 5\% & 10\% & Full & 1\% & 5\% & 10\% & Full & 1\% & 5\% & 10\% & Full \\
\midrule
\multirow{20}{*}{\rotatebox[origin=c]{90}{\textbf{Llama3.1-8B-Instruct}}} 
 & \multirow{4}{*}{Zero-shot} 
   & \cellcolor{gray!10}Test & \multicolumn{4}{c}{\cellcolor{gray!10}36.22}& \multicolumn{4}{c}{\cellcolor{gray!10}16.66}& \multicolumn{4}{c}{\cellcolor{gray!10}27.53} \\
  & & \small{MMLU}   & \multicolumn{4}{c}{--}    & \multicolumn{4}{c}{68.51} & \multicolumn{4}{c}{--}    \\
 & & \small{BBH}   & \multicolumn{4}{c}{--}    & \multicolumn{4}{c}{69.90} & \multicolumn{4}{c}{--}    \\
 & & \small{TyDiQA}   & \multicolumn{4}{c}{--}    & \multicolumn{4}{c}{67.37} & \multicolumn{4}{c}{--}    \\
 
 & \multirow{4}{4cm}{$\textsc{Ngft}_{\text{DS}} + \text{FPFT}$} 
   & \cellcolor{gray!10}Test & \cellcolor{gray!10}63.50& \cellcolor{gray!10}72.32& \cellcolor{gray!10}73.26& \cellcolor{gray!10}77.71& \cellcolor{gray!10}34.74& \cellcolor{gray!10}35.62& \cellcolor{gray!10}36.64& \cellcolor{gray!10}38.10& \cellcolor{gray!10}34.02& \cellcolor{gray!10}36.96& \cellcolor{gray!10}37.76& \cellcolor{gray!10}40.57 \\
 & & \small{MMLU}   & 62.93 & 63.20 & 63.35 & 63.23 & 61.38 & 57.80 & 55.46 & 54.13 & 58.72 & 57.92 & 57.52 & 50.32 \\
\bottomrule
\end{tabular}}
\caption{Comparison of performance improvements on the Llama3.1-8B-Instruct model using different \textsc{Ngft} components individually and in combination.}
\label{tab:NGFT_8B_Llama}
\end{table*}

\end{document}